\documentclass[9.5pt,journal,compsoc]{IEEEtran}

\usepackage{xcolor,soul,framed} 

\colorlet{shadecolor}{yellow}
\usepackage[pdftex]{graphicx}
\graphicspath{{../pdf/}{../jpeg/}}
\DeclareGraphicsExtensions{.pdf,.jpeg,.png}

\usepackage[cmex10]{amsmath}
\usepackage{array}
\usepackage{mdwmath}
\usepackage{mdwtab}
\usepackage{eqparbox}
\usepackage{url}
\usepackage{graphicx}
\usepackage{amsmath}
\usepackage{amsfonts}
\usepackage{algorithm}
\usepackage{algpseudocode}
\usepackage{multirow}
\usepackage{footnote}
\usepackage{bm}
\usepackage{makecell}
\usepackage{pdflscape}
\usepackage{afterpage}
\usepackage{bbding}
\usepackage{lscape}
\usepackage{booktabs}
\usepackage{tikz}
\usepackage{tabularx}
\usepackage{mathrsfs}
\usepackage[figuresright]{rotating}

\usepackage{pgfplots}
\pgfplotsset{compat=1.12}
    \pgfplotsset{
        layers/my layer set/.define layer set={
            background,
            main,
            foreground
        }{ },
        set layers=my layer set,
    }

\usepackage[square,sort,comma,numbers]{natbib} 
\usepackage[top=0.4in,bottom=0.4in,left=0.4in,textwidth=7.7in]{geometry}

\usepackage[pagebackref=false,breaklinks=false,linkcolor=red,anchorcolor=blue, citecolor=green,colorlinks,bookmarks=true]{hyperref}
\usepackage{amsfonts}

\usepackage{pifont}

\begin{document}

\title{Few-shot Class-incremental Learning for Classification and Object Detection: A Survey}
 
\author{\IEEEauthorblockN{Jinghua Zhang\thanks{J. Zhang (zhangjingh@foxmail.com) and Dewen Hu (dwhu@nudt.edu.cn) 
are with the College of Intelligence Science and Technology, National University of Defense Technology (NUDT), Changsha, China. Jinghua Zhang is also with the Center for Machine Vision and Signal Analysis (CMVS), University of Oulu, Finland. Li Liu (dreamliu2010@gmail.com) is with the College of Electronic Science and Technology, NUDT, Changsha, China. Olli Silvén (olli.silven@oulu.fi) and Matti Pietikäinen (matti.pietikainen@oulu.fi) is with CMVS, University of Oulu, Finland.} },
Li Liu,  
Olli Silvén,
Matti Pietikäinen,
Dewen Hu
\thanks{Corresponding authors: Dewen Hu and Li Liu}
\thanks{This work was partially supported by the National Natural Science Foundation of China under grant 62036013 and 62376283, and the Key Stone grant (JS2023-03) of the NUDT.
}
}

\markboth{Accepted by IEEE TPAMI}
{Zhang \MakeLowercase{\textit{et al.}}: }

\IEEEtitleabstractindextext{%
\begin{abstract}
Few-shot Class-Incremental Learning (FSCIL) presents a unique challenge in Machine Learning (ML), as it necessitates the Incremental Learning (IL) of new classes from sparsely labeled training samples without forgetting previous knowledge. While this field has seen recent progress, it remains an active exploration area. This paper aims to provide a comprehensive and systematic review of FSCIL. In our in-depth examination, we delve into various facets of FSCIL, encompassing the problem definition, the discussion of the primary challenges of unreliable empirical risk minimization and the stability-plasticity dilemma, general schemes, and relevant problems of IL and Few-shot Learning (FSL). Besides, we offer an overview of benchmark datasets and evaluation metrics. Furthermore, we introduce the Few-shot Class-incremental Classification (FSCIC) methods from data-based, structure-based, and optimization-based approaches and the Few-shot Class-incremental Object Detection (FSCIOD) methods from anchor-free and anchor-based approaches. Beyond these, we present several promising research directions within FSCIL that merit further investigation.

\end{abstract}

}

\maketitle
\IEEEdisplaynontitleabstractindextext
\IEEEpeerreviewmaketitle

\IEEEraisesectionheading{\section{Introduction}
\label{sec:intro}}
Over the last decade, Deep Neural Networks (DNNs) have gone through several distinct developmental phases: from architectural engineering based on supervised learning as demonstrated by AlexNet~\cite{krizhevsky2012imagenet} and ResNet~\cite{he2016deep}, to the combined strategy of supervised pre-training and fine-tuning, with Transformer-based BERT~\cite{devlin2018bert} being a prime example. This progress further extended to a fusion of self-supervised or semi-supervised pre-training with prompt engineering, as demonstrated by the GPT series~\cite{brown2020language}. These advancements have consistently expanded algorithmic performance boundaries and opened up new application possibilities. However, it's essential to recognize that these DNN achievements have heavily relied on a huge amount of high-quality data, expensive computing hardware, and excellent DNN architectures that are costly to obtain.

DNN learning paradigms are primarily designed for static tasks within a closed-world setting, and it has inherent limitations. Firstly, these models cannot retain previously acquired knowledge and learn new knowledge over time. Specifically, once they are trained on a particular dataset, they often require retraining from scratch when confronted with new tasks or data distributions. Additionally, the process of retraining involves storing vast amounts of old data and updating models, leading to additional computational and storage costs. Such a learning paradigm has at least the following major issues:

\begin{itemize}
  \item Capability and Application Limitations: These systems are optimized for specific tasks they've been trained on, making them ill-suited for dynamic situations. 
  \item Purely Data-Driven Gap: Unlike humans, who learn efficiently with few examples and exhibit lifelong adaptability, these systems rely heavily on vast data and lack the versatility and retention inherent to human learning.
  \item Efficiency and Sustainability Issues: These data and energy-intensive systems require frequent retraining for new data or tasks, increasing computational resource strain and carbon footprint.
  \item Privacy and Security Concerns: The dynamic world exposes these systems to heightened security risks in novel scenarios. Moreover, retaining heightens the risk of data breaches, raising privacy alarms.
\end{itemize}

\begin{figure}[htbp!]
\centering
\includegraphics[width=0.45\textwidth]{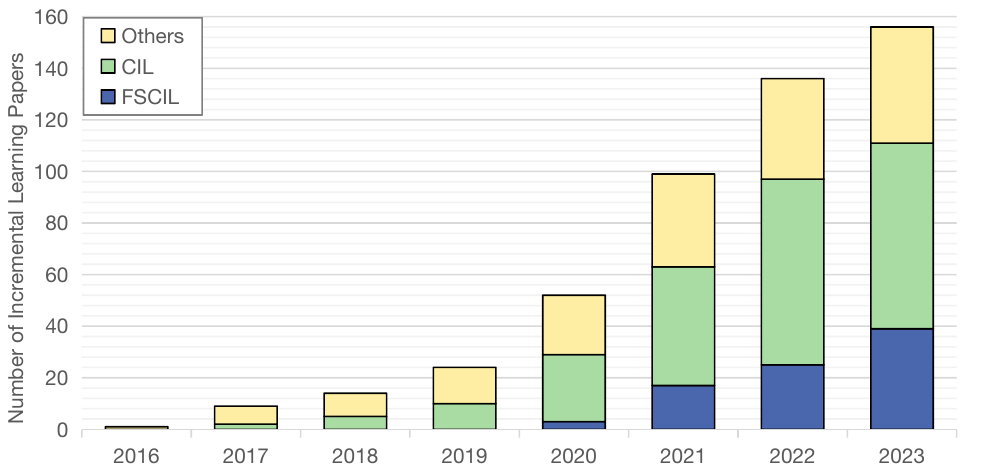}
\caption{IL publications from 2016 to 2023. It is observed that CIL research has become predominant in the field of IL over time, due to its practical value. Concurrently, FSCIL shows a steady rise, mirroring the growing requirement of CIL with limited data.}
\label{Fig1}
\end{figure}

IL, also termed continual or lifelong learning, enables systems to learn new tasks over time while maintaining previous knowledge \cite{9349197,parisi2019continual,9915459}, aiming to replicate human learning abilities \cite{9349197}. This field has seen growing interest recently, prompting numerous studies and surveys \cite{li2023crnet,wu2019large,wang2023cba, 9349197,van2022three,9915459}. The development trend in IL is summarized by the count of academic papers from major conferences and journals, as shown in our collection and the \emph{Awesome-Incremental-Learning} resource\footnote[1]{\url{https://github.com/xialeiliu/Awesome-Incremental-Learning\#2023}}, and depicted in Fig.~\ref{Fig1}. Class-incremental Learning (CIL) is notably prominent, addressing key challenges in real-world scenarios where models should adapt to new classes without forgetting existing ones.

As an important subset of CIL, FSCIL has experienced significant growth over the past four years, shown in Fig.~\ref{Fig1}. It is specifically designed to address the challenges of learning new classes with limited data. This learning paradigm demands that the model retains previously acquired knowledge while continually incorporating new classes, all while dealing with the constraints of limited annotated samples for each class~\cite{ren2019incremental, tao2020few, zhang2021few}. Unlike conventional CIL, FSCIL faces more complex challenges, such as preventing catastrophic forgetting and mitigating overfitting due to sample scarcity. FSCIL seeks to emulate human learning efficiency with minimal data and maintain knowledge over time, making it highly relevant for real-world settings with limited, evolving data. To highlight its practical importance, we provide a concise summary of FSCIL's practical significance:

\begin{itemize}
  \item Adaptation to Dynamic World: FSCIL empowers models to acquire new classes while retaining previous knowledge, a critical capability for effectively adapting to a dynamically changing world.
  \item High Data Efficiency: FSCIL can mitigate the necessity for extensive sample labeling, providing advantages in situations with limited data and high labeling costs.
  \item Environmental Sustainability: FSCIL promotes sustainability by requiring fewer computational and storage resources than traditional methods, a crucial benefit in resource-limited environments.
  \item Data Security and Privacy: FSCIL reduces the need to retain extensive historical data, thereby aligning with data security and privacy requirements.
  \item Versatile Applications: FSCIL is applicable in various fields, especially where data is limited, labeling is costly, and frequent class updates are needed
\end{itemize}

\begin{figure*}[htb]
\centering
\includegraphics[width=0.95\textwidth]{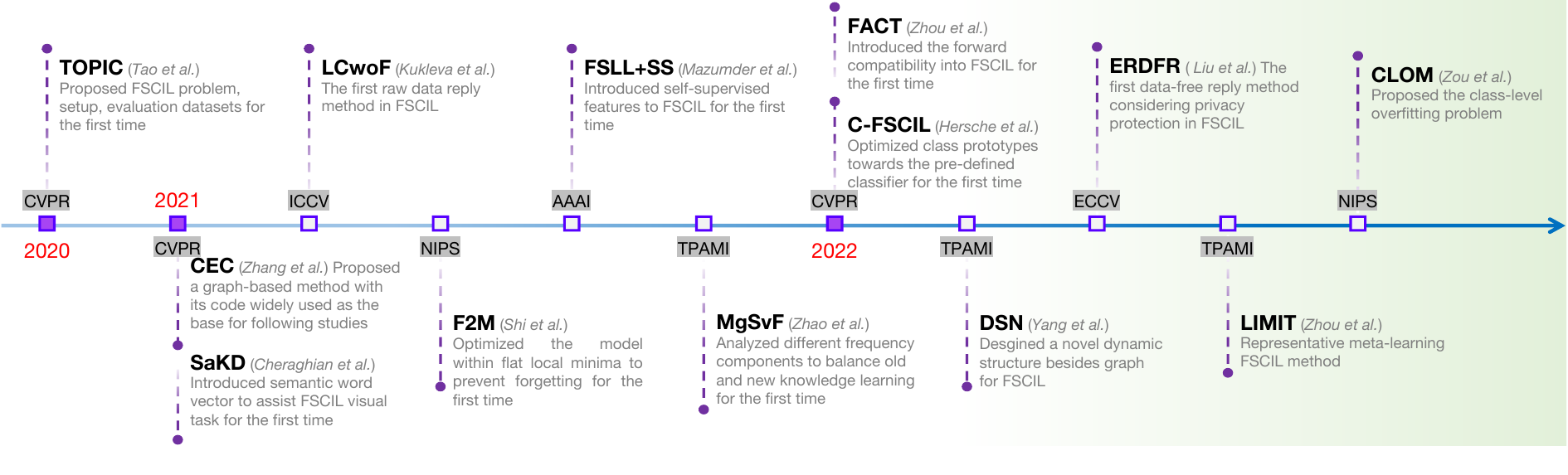}
\caption{A chronological overview of some representative FSCIL methods.
FSCIL was first carried out by TOPIC~\cite{tao2020few}. CEC~\cite{zhang2021few} was widely used as a base for subsequent studies. SaKD~\cite{cheraghian2021semantic} integrated semantic word vectors into FSCIL, offering a reference for applying language-image models in the future. LCwoF~\cite{kukleva2021generalized} and ERDFR~\cite{liu2022few} proposed distinct Data Replay (DR) strategies. F2M~\cite{shi2021overcoming} introduced a novel approach by constraining optimization within flat local minima. FSLL+SS~\cite{mazumder2021few} introduced the semi-supervised features to FSCIL for the first time. MgSvF~\cite{zhao2021mgsvf} analyzed and utilized different frequency components to balance the old and new knowledge. FACT~\cite{zhou2022forward} introduced a fresh perspective by advocating forward compatibility in FSCIL. C-FSCIL~\cite{hersche2022constrained} pre-defined classifiers to guide model optimization. DSN~\cite{yang2022dynamic} offered a novel dynamic structure for FSCIL. LIMIT~\cite{zhou2022few} proposed a representative meta-learning paradigm for FSCIL. CLOM~\cite{zou2022margin} pointed out the issue of class-level overfitting made by metric learning in FSCIL.}
\label{Fig2}
\end{figure*}

Although there has been some progress in the field of FSCIL and some representative works~\cite{gidaris2018dynamic,tao2020few,zhang2021few,9349197,zhou2022few} have emerged, it is yet in its development stage. The key milestones from 2020 to the present are illustrated in Fig.~\ref{Fig2}. Current methods still have a gap to meet the practical applications. Therefore, it is imperative to systematically review the latest developments in this field, identify the core challenges and open questions that hinder its development, and determine the promising future direction. Nevertheless, most of the research on FSCIL is still quite dispersed, and this field needs a systematic and comprehensive survey. It has inspired our survey, which aims to fill the gap. Since it is an ML problem proposed in the field of computer vision in recent years and most of the research work is based on the deep learning algorithm, the scope discussed in our paper is mainly the deep FSCIL algorithm in the field of computer vision, which includes primarily classification and object detection tasks.

Despite existing surveys on FSL~\cite{wang2020generalizing,lu2023survey,antonelli2022few,huang2022survey} and IL~\cite{parisi2019continual,lesort2020continual,9349197,belouadah2021comprehensive,9915459,mai2022online,van2022three,zhou2024class,wang2024comprehensive}, there is a clear lack of systematic and comprehensive surveys specifically on FSCIL. Existing FSL surveys primarily focus on tackling ML problems with limited data. While they provide systematic classifications of FSL methods, they do not touch upon the issue of FSCIL. Similarly, IL surveys mostly focus on ML problems in a continual learning scenario. For instance, \emph{De Lange et al.}~\cite{9915459} conducted a systematic review of DR, regularization, and parameter isolation, along with comparative experiments, while \emph{Zhou et al.}~\cite{zhou2024class} summarized the CIL problem from perspectives like DR, Data Regularization, and Dynamic Networks. However, none of these works systematically review FSCIL. Although some surveys briefly mention FSCIL~\cite{belouadah2021comprehensive,wang2024comprehensive}, they only provide a short introduction to the concept and a few studies without offering a thorough or systematic analysis. Although \emph{Tian et al.}~\cite{tian2024survey} recently conducted a review of FSCIL, its introduction to FSCIL is not in-depth and comprehensive enough.

In this regard, we summarize existing surveys in Tab.~\ref{tab1} and systematically describe the uniqueness of our paper to highlight its unique contributions. To address the shortcomings in FSCIL research, we systematically summarize the field from various aspects, including definition, challenges, general schemes, related problems, datasets, metrics, methods, performance comparisons, and future directions. Our contributions are:

\begin{itemize}
\item Our survey offers a systematic and comprehensive review of classification and object detection methods in FSCIL.
\item We cover problem definition, core challenges, general schemes, related ML problems, benchmark datasets, and evaluation metrics in detail.
\item A structured taxonomy is offered for FSCIL, discussing classification methods from data, structure, and optimization perspectives, and detection methods from anchor-based and anchor-free perspectives.
\item Valuable insights and outlooks in FSCIL are discussed.
\end{itemize}

\begin{table*}[htbp!]
\scriptsize
\centering
\renewcommand{\arraystretch}{1.05}
\setlength\tabcolsep{4pt}
\caption{A summary and comparison of the primary surveys in the fields of FSL, IL, and FSCIL.} 
\vspace{-10pt}
\begin{tabular}{|c|c|c|l|l|}
\hline
\textbf{Topic}          & \textbf{Year} & \textbf{Venue}     & \multicolumn{1}{c|}{\textbf{Title}}                                                                       & \multicolumn{1}{c|}{\textbf{Main Content}}                                                                                                                                                                                                                                                                                                     \\ \hline
\multirow{4}{*}{FSL}   & 2020          & ACM CS             & \begin{tabular}[c]{@{}p{5.4cm}@{}}Generalizing from a Few Examples: A Survey on Few-shot Learning~\cite{wang2020generalizing}\end{tabular}                                           & \begin{tabular}[c]{@{}p{9.5cm}@{}}This paper clearly introduces FSL, including its definition, challenges, and main methods from data, model, and algorithm perspectives. However, it lacks a summary of datasets and performance.\end{tabular}                                                                                                                                       \\ \cline{2-5} 

                       & 2022          & ACM SC             & \begin{tabular}[c]{@{}p{5.4cm}@{}}Few-Shot Object Detection: A Survey~\cite{antonelli2022few}\end{tabular}                                                                       & \begin{tabular}[c]{@{}p{9.5cm}@{}}This paper focuses on few-shot object detection. It reviews related methods based on data augmentation, transfer learning, metric learning, and meta-learning. However, it lacks a summary of challenges and relevant problems.\end{tabular}                                                                                                                   \\ \cline{2-5} 
                       & 2022          & TPAMI              &

                       \begin{tabular}[c]{@{}p{5cm}@{}}A Survey of Self-Supervised and Few-Shot Object Detection~\cite{huang2022survey}\end{tabular}                                                 & \begin{tabular}[c]{@{}p{9.5cm}@{}}This paper reviews few-shot object detection methods in self-supervised and few-shot object detection, highlighting the potential of technique fusion. However, it lacks a summary of challenges and relevant problems.\end{tabular}     \\ \cline{2-5}
                       
                      & 2023          & PR              & \begin{tabular}[c]{@{}p{5.4cm}@{}}A survey on machine learning from few samples~\cite{lu2023survey}\end{tabular}                                                                 & \begin{tabular}[c]{@{}p{9.5cm}@{}}This paper categorizes FSL approaches into generative and discriminative models. It also discusses emerging FSL topics and its various applications. However, it lacks a summary of challenges and relevant problems.\end{tabular}                                                                                                                             \\  
                       
                       \hline
\multirow{8}{*}{IL}    & 2019          &  \begin{tabular}[c]{@{}c@{}}Neural \\Networks \end{tabular}  & \begin{tabular}[c]{@{}p{5cm}@{}}Continual Lifelong Learning with Neural Networks: A Review~\cite{parisi2019continual}\end{tabular}                                                & \begin{tabular}[c]{@{}p{9.5cm}@{}}This paper discusses the challenges and approaches in lifelong learning, which involves continuous knowledge acquisition and application, and emphasizes the need to address catastrophic forgetting. But it lacks a summary of relevant problems and performance.\end{tabular}                                                                           \\ \cline{2-5} 
                       & 2020          &  \begin{tabular}[c]{@{}c@{}}Information \\ Fusion \end{tabular} & \begin{tabular}[c]{@{}p{5.2cm}@{}}Continual Learning for Robotics: Definition, Framework, Learning Strategies, Opportunities and Challenges~\cite{lesort2020continual}\end{tabular} & \begin{tabular}[c]{@{}p{9.5cm}@{}}This paper reviews continual learning, highlights the need for stable real-world algorithms, summarizes existing benchmarks and metrics, and proposes a framework for evaluating approaches across both robotics and non-robotics fields. However, it lacks a summary of performance.\end{tabular}                                                             \\ \cline{2-5} 
                       & 2021          & \begin{tabular}[c]{@{}c@{}}Neural \\Networks \end{tabular}   & \begin{tabular}[c]{@{}p{5.4cm}@{}}A Comprehensive Study of Class Incremental \\Learning Algorithms for Visual Tasks~\cite{belouadah2021comprehensive} \end{tabular}                           & \begin{tabular}[c]{@{}p{9.5cm}@{}}This paper focuses on CIL, providing the taxonomy of regularization approaches, dynamic architectures, and complementary learning systems and memory replay. However, it lacks a summary of challenges and relevant problems.\end{tabular}                                                                                              \\ \cline{2-5} 
                       & 2022          & TPAMI              & \begin{tabular}[c]{@{}p{5.4cm}@{}}A Continual Learning Survey: Defying Forgetting in Classification Tasks~\cite{9349197}\end{tabular}                                   & \begin{tabular}[c]{@{}p{9.5cm}@{}}This paper summarizes Task-incremental Learning (TIL) and compares 11 advanced methods. It also analyzes the impact of model parameters, task order, and resource requirements. However, it lacks a summary of challenges and relevant problems.\end{tabular}                                                                                                        \\ \cline{2-5} 
                       & 2022          & TPAMI              & \begin{tabular}[c]{@{}p{5.4cm}@{}}Class-Incremental Learning: Survey and Performance Evaluation on Image Classification~\cite{9915459}\end{tabular}                     & \begin{tabular}[c]{@{}p{9.5cm}@{}}This paper surveys CIL methods for image classification. It provides experimental evaluations on 13 methods and explores various scenarios. However, it lacks a summary of relevant problems.\end{tabular}                                                                                                                              \\ \cline{2-5} 
                       & 2022          & \begin{tabular}[c]{@{}c@{}}Neuro- \\ Computing \end{tabular}
                        & \begin{tabular}[c]{@{}p{5.4cm}@{}}Online Continual Learning in Image Classification: An Empirical Survey~\cite{mai2022online} \end{tabular}                                    & \begin{tabular}[c]{@{}p{9.5cm}@{}}This study reviews online continual learning in image classification and compares advanced methods. However, it lacks a summary of challenges and relevant problems.\end{tabular}                                                                                                                                                                               \\ \cline{2-5} 
                       & 2024          & TPAMI              & \begin{tabular}[c]{@{}p{5.4cm}@{}}A Comprehensive Survey of Continual Learning:\\ Theory, Method and Application~\cite{wang2024comprehensive}\end{tabular}                              & \begin{tabular}[c]{@{}p{10cm}@{}}This survey explores continual learning from theoretical foundations, methods, and applications, but it lacks a summary of relevant problems, datasets, and performance.\end{tabular}                                                                                                                                                             \\ \cline{2-5} 
                       & 2024          & TPAMI              & \begin{tabular}[c]{@{}p{5.4cm}@{}} Class-Incremental Learning: A Survey~\cite{zhou2024class}\end{tabular}                                                                 & \begin{tabular}[c]{@{}p{10cm}@{}}This paper categorizes methods into data-centric, model-centric, and algorithm-centric, evaluating 16 methods and advocating for fair comparisons. However, it lacks a summary of challenges and relevant problems.\end{tabular}                                                                                         \\                                                                                    \hline
\multirow{2}{*}{FSCIL} & 2023          & \begin{tabular}[c]{@{}c@{}}Neural \\Networks \end{tabular}              & \begin{tabular}[c]{@{}p{5.4cm}@{}}A Survey on Few-Shot Class-Incremental \\Learning~\cite{tian2024survey}\end{tabular}                                                           & \begin{tabular}[c]{@{}p{9.5cm}@{}}This survey focuses on FSCIL. However, the taxonomy and summary of relevant problems are not systematic. It also lacks a summary of the challenges. Discussion about object detection and the future direction is not in-depth. \end{tabular}                                     \\ \cline{2-5} 
                       & \textbf{2023} & \textbf{Ours}      & \begin{tabular}[c]{@{}p{5.4cm}@{}}\textbf{Few-shot Class-incremental Learning: A Survey}\end{tabular}                                                    & \begin{tabular}[c]{@{}p{9.5cm}@{}}\textbf{Our paper focuses on FSCIL in classification and object detection tasks, clarifies the challenge and relationship with relevant problems, and provides a systemical taxonomy from data-based, structure-based, and optimization-based approaches. We also summarize the performance of existing methods and point out potential directions.}\end{tabular} \\ \hline
\end{tabular}
\label{tab1}
\end{table*}

The paper structure is as follows: Sec.~\ref{sec:overv} presents a detailed overview of FSCIL, including its definition, challenges, general frameworks, and its relationship with relevant problems. Sec.~\ref{sec:DE} discusses popular FSCIL datasets and evaluation metrics. Sec.~\ref{sec:FSCIC} examines FSCIC methods from data, structure, and optimization perspectives, and Sec.~\ref{sec:FSCIOD} covers FSCIOD methods from anchor-based and anchor-free viewpoints. The paper concludes in Sec.~\ref{sec:CO} with a summary and future directions.

\section{Background}
\label{sec:overv}

\subsection{Problem Definition}
\label{sec:over:pd}

\begin{figure*}[htbp!]
\centering
\includegraphics[width=0.95\textwidth]{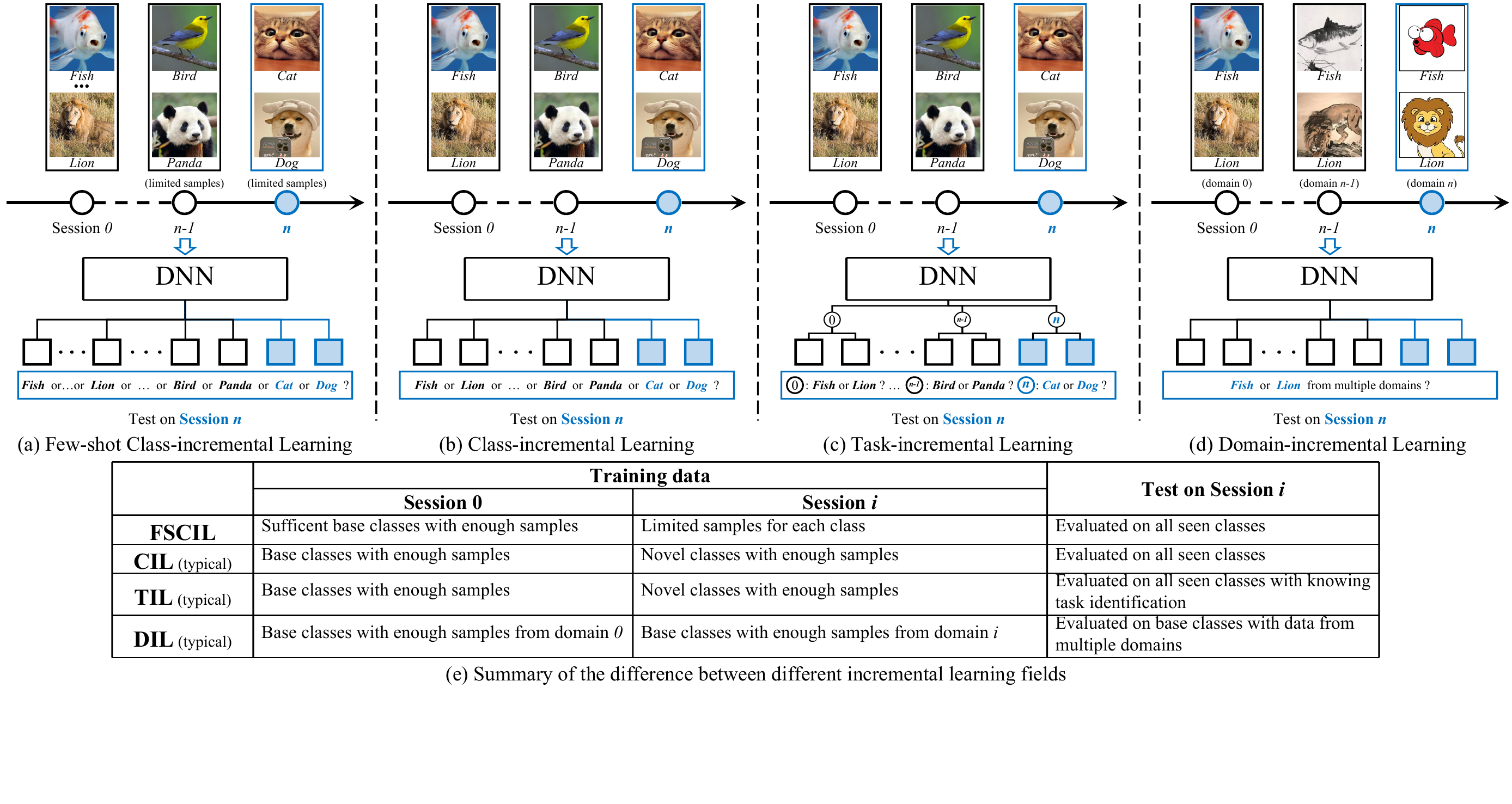}
\caption{The general settings of different IL tasks. Specifically, (a) shows the setting of FSCIL, (b) is the CIL, (c) represents the setting of TIL, and (d) illustrates the Domain-incremental Learning (DIL). FSCIL can be viewed as a subdomain of CIL, where the base session usually has sufficient training data, and the incremental sessions are formed in the $N-$way $K-$shot format. TIL differs from CIL because the session identity is known during model training and testing. In contrast, DIL maintains the same classification tasks, but the data across different sessions comes from different domains. Note that ``session'' may also be called ``task'' in other literature.}
\label{Fig3}
\end{figure*}

The FSCIL aims to learn an ML model that can continuously learn knowledge from a sequence of new classes with only a few labeled training samples while preserving the knowledge gained from previous classes~\cite{tao2020few,zhang2021few,ji2023memorizing,wang2023few}. Taking the classification task as an example, Fig.~\ref{Fig3}(a) offers an overview of the general setting for FSCIL, including the setting of training data, the model learning process, and the evaluation setting. 

\textbf{Setting:} As shown in Fig.~\ref{Fig3}(a), the data stream used in FSCIL contains a base session and a sequence of new sessions. The training datasets in these sessions can be denoted by $\{D^0_{train},D^1_{train},\cdots,D^B_{train}\}$, where $B$ is the number of new sessions. The base training dataset generally contains sufficient labeled samples from the distribution $D^0_t$, and it can be formulated by $D^0_{train}=\{(x_i,y_i)\}^{n_0}_{i=1}$, where $n_0$ is the number of training samples in the base session, $x_i$ is a training sample from class $y_i \in Y_0$, and $Y_0$ is the corresponding label space of $D^0_{train}$. Differently, the training dataset in each new session is in the form of $N-$way $K-$shot, where $N-$way means that the training set contains $N$ classes and $K-$shot means each class contains $K$ labeled samples. It can be formulated as $\forall$ integer $b\in[1,B], D^b_{train}=\{(x_i,y_i)\}^{N\times K}_{i=1}$. Note that the classes in different sessions do not intersect, \emph{i.e.}, $\forall$ integer $p,q\in[0,B]$ and $p\not=q$, $Y_p {\cap} Y_q = \emptyset$.

\textbf{Model:} During the training session $b$, the dataset $D^b_{train}$ is accessible, and the original complete training datasets from previous sessions are unavailable (note: some methods like DR or KD may store a few old samples to revisit old knowledge). The FSCIL model must learn new classes from $D^b_{train}$ while maintaining performance on old classes, \emph{i.e.}, minimizing the expected risk $\mathcal{R}\left(f,b\right)$ on all the seen classes~\cite{zhou2022few,9349197}. This process is formulated as follows:
\begin{equation}
\label{eq1}
\begin{aligned}
\mathbb{E}_{(x_i,y_i)\sim D^0_{t} \cup \cdots  D^b_{t}}\left[ L \left( f\left( x_i; \mathcal{D}^b_{train},{\boldsymbol{\theta}^{b-1}} \right), y_i\right) \right],
\end{aligned}
\end{equation}
where the current FSCIL algorithm $f$ aims to built the new model based on the dataset $\mathcal{D}^b_{train}$ and previous model parameters ${\boldsymbol{\theta}^{b-1}}$ and minimize the loss $L$ on all seen classes~\cite{zhou2022few}. Because the datasets in FSCIL are continuously updated, the expected risk on every new session should be optimized, \emph{i.e.}, $\sum^B_{b=1}\mathcal{R}\left(f,b\right)$ should be optimized.

\textbf{Evaluation:} The testing datasets in FSCIL sessions can be denoted by $\{D^0_{test},\cdots,D^B_{test}\}$, which shares the same label space as their corresponding training datasets. For the evaluation in session $b$, the FSCIL model needs to be evaluated by the joint testing datasets, which encompass all the testing datasets from the current and all preceding sessions, denoted as $D^0_{test} \cup \cdots \cup D^b_{test}$. This measure helps quantify the model's performance across all classes it has encountered up to that point.

\subsection{Core Challenges}
\label{sec:over:cc}
FSCIL faces significant challenges, notably the unreliable empirical risk minimization and the stability-plasticity dilemma. In FSCIL sessions, limited supervised data mean empirical risk fails to accurately represent expected risk, decreasing model generalization and increasing overfitting risks. Moreover, as new classes are continually added, old knowledge can be easily forgotten and overwritten by new knowledge. This leads to catastrophic forgetting. Otherwise, intransigence may occur. Therefore, balancing model stability and plasticity is another core challenge. This section provides the details of these challenges.

\subsubsection{Unreliable Empirical Risk Minimization}
In FSCIL, unreliable empirical risk minimization, where the model is trained to minimize prediction errors on the training data, poses a major challenge. This approach doesn't ensure strong generalization on test data, especially with limited training samples. In FSCIL, each session's training dataset follows an $N-$way $K-$shot format, often leading to a significant discrepancy between empirical and expected risks due to inadequate samples for new classes. This gap can result in overfitting, where the model excels on training data but underperforms on testing data, compromising its generalization ability~\cite{wang2020generalizing,song2023comprehensive}.

In contrast to conventional FSL, FSCIL not only grapples with the issue of scarce samples but is also confronted with the challenge posed by the continual increase in classes. Continuous unreliable empirical risk minimization in successive sessions may hinder the model's convergence to an ideal state, questioning not only the reliability of the model formed in the current incremental session but also presenting a challenge in maintaining model stability in the subsequent incremental session. This issue becomes particularly pronounced when dealing with multiple incremental classes with limited training samples~\cite{shi2021overcoming}.

To elaborate on this challenge, we introduce essential concepts of empirical risk minimization ~\cite{wang2020generalizing,bottou2007tradeoffs,bottou2018optimization}. For a learning task with dataset $D = \{D_{train}, D_{test}\}$, where $p(x, y)$ denotes the joint probability distribution of data $x$ and label $y$, and $f_o$ is the optimal hypothesis from $x$ to $y$, \emph{i.e.}, the function that minimizes the expected risk. Specifically, given a hypothesis $f$, the expected risk $\mathcal{R}\left(f\right)$, which measures the loss concerning $p(x, y)$, is formulated as: 
\begin{equation}
\label{eq2}
\begin{aligned}
\mathcal{R}\left(f\right)=\int L \left( f\left( x \right), y\right)dp\left(x,y\right) =\mathbb{E}\left[ L \left( f\left( x \right), y\right) \right],
\end{aligned}
\end{equation}
and $f_o$ can be explained as:
\begin{equation}
\label{eq3}
f_o= \mathop{\arg\min}_{f}\mathcal{R}\left(f\right).
\end{equation}
As $p(x, y)$ is unknown, the empirical risk, which is the average loss value obtained on the training dataset $D_{train}$ of $I$ samples, is generally used as a proxy of $\mathcal{R}(f)$ for minimization. Specifically, empirical risk can be formulated as:
\begin{equation}
\label{eq4}
\begin{aligned}
\mathcal{R}_I\left(f\right)= \frac{1}{I}\sum^{I}_{i=1} L \left( f\left( x \right), y\right).
\end{aligned}
\end{equation}
Since $D_{train}$ is deterministic, a hypothesis space $\mathcal{F}$ of hypotheses $f(\boldsymbol{\theta})$ is chosen to optimize the model. The minimization of $\mathcal{R}_I(f)$ can be denoted as:
\begin{equation}
\label{eq5}
f_e= \mathop{\arg\min}_{f\in\mathcal{F}}\mathcal{R}_I\left(f\right).
\end{equation}
Ideally, $f_e$ approximates $f_o$ as closely as possible. However, since $f_o$ is unknown, it requires some $f \in \mathcal{F}$ to approximate it. Assume $f_b$ is the best approximation for $f_o$ in $\mathcal{F}$, which can be formulated as:
\begin{equation}
\label{eq6}
f_b= \mathop{\arg\min}_{f\in\mathcal{F}}\mathcal{R}\left(f\right).
\end{equation}
Eclectically, we hope $f_e$ can approximate $f_b$ as closely as possible. For simplicity, we assume that $f_o$, $f_e$, and $f_b$ are well-defined and unique. The total error can be decomposed as:
\begin{equation}
\label{eq7}
\mathbb{E}\left[\mathcal{R}\left(f_e\right)-\mathcal{R}\left(f_o\right)\right] = \underbrace{\mathbb{E}\left[\mathcal{R}\left(f_b\right)-\mathcal{R}\left(f_o\right)\right]}_{\mathcal{E}_{app}} + \underbrace{\mathbb{E}\left[\mathcal{R}\left(f_e\right)-\mathcal{R}\left(f_b\right)\right]}_{\mathcal{E}_{est}}.
\end{equation}
Here, the expectation concerns the random choice of $D_{train}$. The approximation error $\mathcal{E}_{app}$ measures how closely functions in $\mathcal{F}$ can approximate the optimal hypothesis $f_o$, and the estimation error $\mathcal{E}_{est}$ measures the effect of minimizing the empirical risk $\mathcal{R}_I(f)$ instead of the expected risk $\mathcal{R}(f)$ in $\mathcal{F}$. Overall, the hypothesis space $\mathcal{F}$ and the number of examples in $D_{train}$ affect the total error~\cite{wang2020generalizing}.

\begin{figure}[htbp!]
\centering
\includegraphics[width=0.37\textwidth]{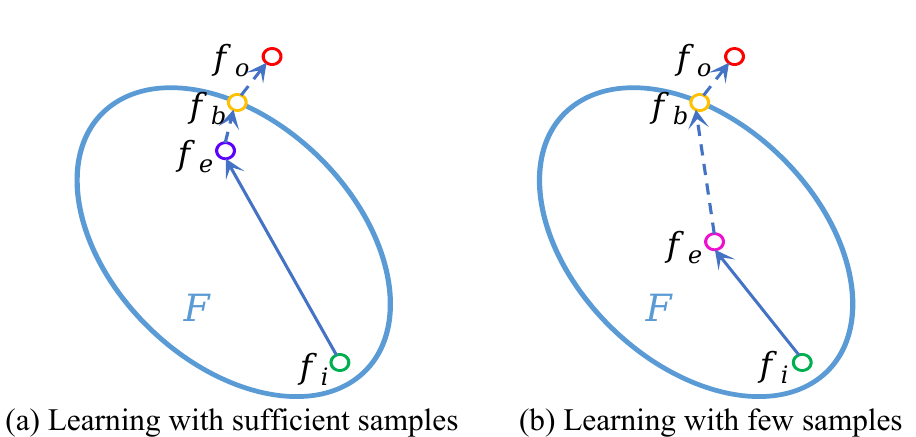}
\caption{The illustration of unreliable empirical risk minimization in FSCIL. (a) with sufficient training samples, the empirical risk minimization can approximate the best-expected risk minimization function. (b) when the training samples are insufficient, the best empirical risk minimization function is often a poor approximation to the best-expected risk minimization function.}
\label{Fig4}
\end{figure}

As illustrated in Fig.~\ref{Fig4}(a), when the supervised information in $D_{train}$ is sufficient, i.e., $I$ in $D_{train}$ is large enough, the empirical risk minimization function in $\mathcal{F}$ can approximate the best-expected risk minimization function in $\mathcal{F}$ well, i.e., $f_e$ can provide a good approximation to $f_b$. However, due to the limited number of training samples in each FSCIL incremental session, the best empirical risk minimization function is often a poor approximation to the best-expected risk minimization function in $\mathcal{F}$, i.e., $f_e$ is far from $f_b$ in $\mathcal{F}$, as shown in Fig.~\ref{Fig4}(b). This discrepancy leads to unreliable empirical risk minimization in the model learning process.

\subsubsection{Stability-plasticity Dilemma}
In FSCIL, a central challenge is the stability-plasticity dilemma, which involves balancing the model's consistent performance on learned classes (stability) and its adaptability to new classes with limited samples (plasticity). Traditional deep learning models are typically static and can only handle previously learned classes. FSCIL demands continual learning of new classes with only a few available labeled training samples and without access to the original complete training data of old classes. It requires the model to maintain the stability of previously learned knowledge and plasticity in learning new knowledge. Due to different optimization goals for old and new classes, the decision boundary often shifts toward new classes, leading to catastrophic forgetting. Conversely, focusing too much on old knowledge stability may limit the ability to learn new tasks, a phenomenon known as intransigence. Therefore, balancing stability and plasticity is crucial in FSCIL.

\begin{figure}[htbp!]
\centering
\includegraphics[width=0.37\textwidth]{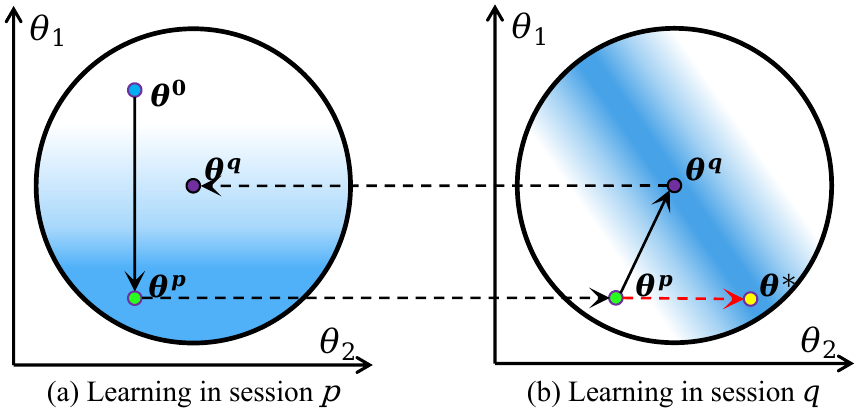}
\caption{The illustration of stability-plasticity dilemma in FSCIL. (a) and (b) are two consecutive sessions. Darker areas indicate optimal loss values. $\boldsymbol{\theta^p}$ performs well in session $p$ but poorly in $q$. Optimizing $\boldsymbol{\theta^p}$ to $\boldsymbol{\theta^q}$ on session $q$ diminishes its performance on session $p$. Yet, directing optimization towards $\boldsymbol{\theta^\star}$ ensures good results on both sessions.}
\label{Fig5}
\end{figure}

The stability-plasticity dilemma can be illustrated through consecutive sessions $p$ and $q$. Fig.~\ref{Fig5}(a) and Fig.~\ref{Fig5}(b) depict error surfaces for these sessions, with darker areas representing ideal loss values, and the model under consideration has only two parameters, $\theta_1$ and $\theta_2$. It can be observed that the optimization objective of session $p$ is to move downwards, while that of session $q$ is to approach the band center. Suppose the initial model on session $p$ is $\boldsymbol{\theta^0}$, and the optimized is $\boldsymbol{\theta^p}$, which shows promising performance on session $p$. However, when the model starts learning the next session $q$, $\boldsymbol{\theta^p}$ obtained from session $p$ is insufficient to meet the requirement of session $q$. To solve the problem, the model usually adjusts the parameters to minimize the loss towards the center of the loss surface. Assuming the optimized model for session $q$ is $\boldsymbol{\theta^q}$, it can be observed that $\boldsymbol{\theta^q}$ can adapt well to the analysis tasks on session $q$. However, when we use $\boldsymbol{\theta^q}$ to make predictions on session $p$, the decision boundary cannot achieve satisfactory performance, indicating the occurrence of forgetting. Nevertheless, if we constrain $\boldsymbol{\theta^p}$ to move towards $\boldsymbol{\theta^\star}$ while learning session $q$, we can observe that the model can adapt to both session $p$ and session $q$ effectively.

To balance model stability and plasticity in a new session, the key approach is distinguishing between critical and non-critical parameters from the previous session, optimizing only the non-critical ones. The loss function for the new session encompasses both the classification task and prevention of catastrophic forgetting. It is formulated as follows:
\begin{equation}
\label{eqxxx}
\begin{aligned}
L^{\prime}\left(\boldsymbol{\theta}\right)=L\left(\boldsymbol{\theta}\right)+\lambda\sum_i b_i\left(\theta_i-\theta^b_i\right)^2,
\end{aligned}
\end{equation}
where $L$ is the partial loss function for the current classification task, $\theta_i$ denotes the parameter in the current model $\boldsymbol{\theta}$, $\theta^b_i$ represents the corresponding parameter in the previous model $\boldsymbol{\theta^b}$, $b_i$ characterizes the importance of $\theta^b_i$ for the previous task, and the hyperparameter $\lambda$ balances the two parts of the overall loss. Setting $b_i=0$ imposes no constraint on $\theta_i$, leading to catastrophic forgetting. Conversely, setting $b_i=\infty$ results in intransigence, where $\theta_i$ always equals $\theta^b_i$.

\begin{figure}[htbp!]
\centering
\includegraphics[width=0.49\textwidth]{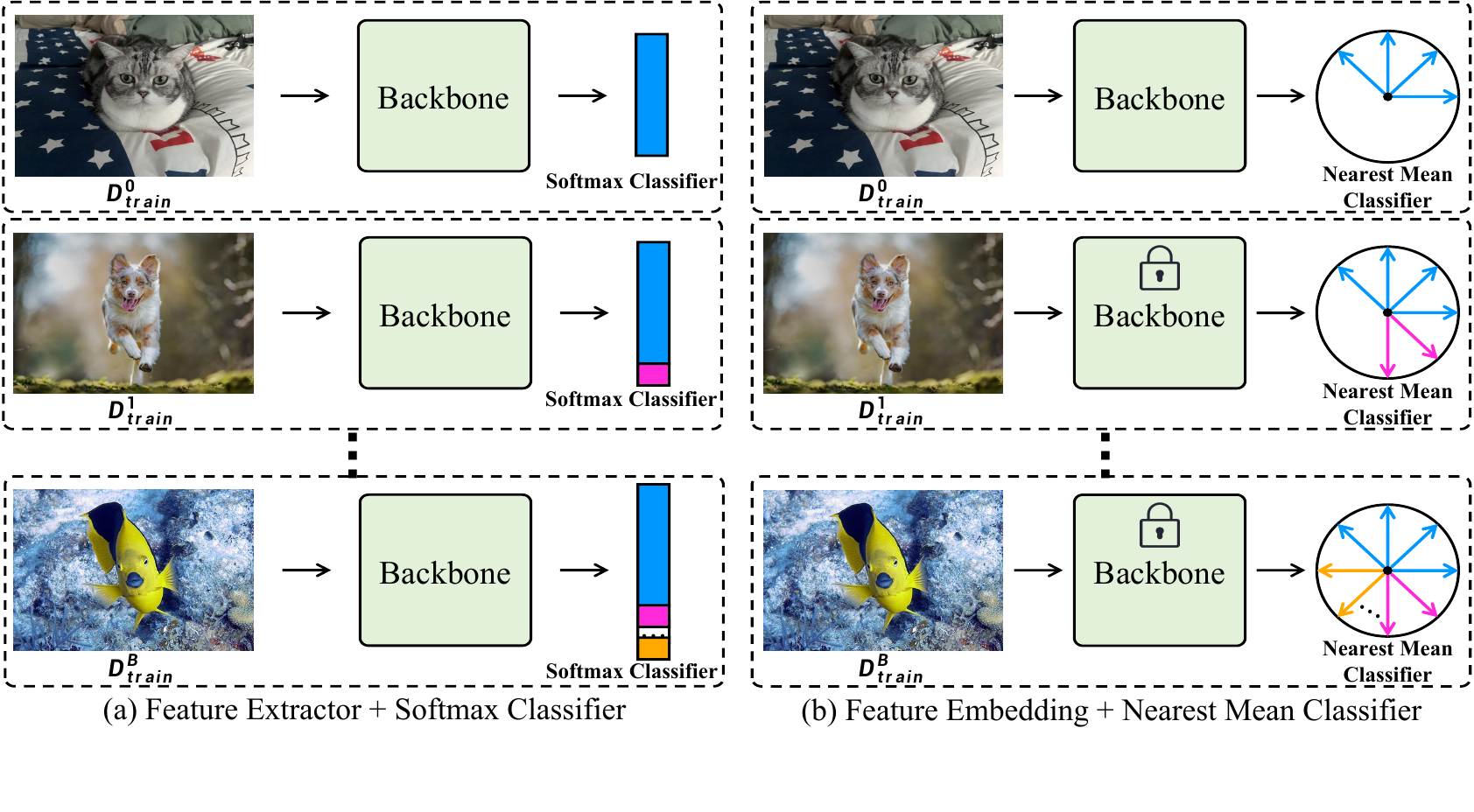}
\caption{The general schemes of FSCIL. (a) ``Feature Extractor + Softmax Classifier'' has a trainable backbone with Knowledge Distillation (KD)-based methods being the most typical method. (b) is ``Feature Embedding + Nearest Mean Classifier'' with a fixed backbone after base training.}
\label{Fig6}
\end{figure}

\subsection{General Schemes}
\label{sec:over:gs}
FSCIL has two main frameworks, as shown in Fig.~\ref{Fig6}. The first uses a feature extractor with a softmax classifier, while the second involves a feature embedding network and the nearest class mean classifier~\cite{zhao2021mgsvf,yu2020semantic,rebuffi2017icarl}. The entire network is trainable throughout the IL process in the first one. To mitigate catastrophic forgetting, some studies~\cite{zhu2022feature,cui2022uncertainty,cui2023uncertainty} use KD to maintain competent classification capabilities on previous classes while learning new ones. The second framework focuses on training a feature embedding network to map samples into a space where distances represent semantic differences, followed by classification using the nearest class mean classifier. For instance, some studies~\cite{peng2022few} employ metric loss for the training of the embedding network, enabling it to learn more discriminative features and better adapt to incremental classes.

\subsection{Relevant Problems}
\label{sec:over:rp}

\subsubsection{Incremental Learning}
This section reviews the relationship and distinctions between FSCIL and other IL scenarios, specifically CIL, TIL, and DIL, as outlined by \emph{Van de Ven et al.}~\cite{van2022three}.

\textbf{Class-incremental Learning:}
CIL aims to learn an algorithm that can continuously recognize new classes without forgetting old ones~\cite{van2022three,9349197,zhou2024class}. As FSCIL can be seen as a subdomain of CIL, it can be observed from Fig.~\ref{Fig3}(a) and Fig.~\ref{Fig3}(b) that their general settings are very similar. Both require learning new class data as it arrives and maintaining classification abilities on previous classes. However, FSCIL's base session often includes many training samples, while CIL does not have strict restrictions. Additionally, the training samples in the incremental session of FSCIL are limited and exist in the form of $N-$way $K-$shot. In contrast, the training samples in the incremental session of CIL are usually sufficient. The core challenge of CIL lies in solving the stability-plasticity dilemma. At the same time, FSCIL needs to solve this challenge and address the problem caused by unreliable empirical risk minimization due to the lack of training samples and its sustained impact in continuous scenarios.

\textbf{Task-incremental Learning:}
TIL aims to learn an algorithm that can progressively learn new tasks without forgetting old ones. As depicted in Fig.\ref{Fig3}(c), TIL's training data in classification scenarios is split into multiple sessions, each representing a distinct task. During both training and testing, the TIL model is always aware of the specific task identity. To avert catastrophic forgetting, various algorithms\cite{mallya2018packnet,maltoni2019continuous,van2022three} employ task-specific components or design separate networks for each task. TIL's primary challenge lies in identifying shared features across tasks to balance performance and computational complexity, using knowledge from one task to enhance performance in others~\cite{van2022three}.

\textbf{Domain-incremental Learning:}
DIL is an ML problem designed to continuously adapt to data distribution from different domains while the structure of the problem is always the same~\cite{van2022three}. DIL addresses the variation in data distribution across incremental domains, enabling effective learning and prediction in new domains without forgetting previously acquired knowledge. As depicted in Figure~\ref{Fig3}(d), DIL involves training data from multiple sessions, each containing identical classes but with distinct data distributions indicative of different domains. The DIL model must continuously adapt to these new domains without losing prior knowledge. Its primary challenge is to identify and leverage shared features across domains, allowing quick adaptation to new domains and learning new knowledge while preserving existing knowledge in old domains.

\subsubsection{Few-shot Learning}
FSL refers to using very few training samples for model learning~\cite{hu2022can}. To better understand the correlations and distinctions between FSCIL and FSL, this section presents pertinent concepts, including FSL and general Few-shot Learning (gFSL). For clarity, Tab.~\ref{tab2} is provided, summarizing the distinct attributes of FSL, gFSL, and FSCIL.

\begin{table}[htb]
\scriptsize
\centering
\renewcommand{\arraystretch}{1}
\setlength\tabcolsep{9pt}
\caption{The difference between FSL, gFSL, and FSCIL. Note that the base classes indicate the original complete version of base training data.} 
\vspace{-10pt}
\begin{tabular}{|c|cc|c|}
\hline
\multirow{2}{*}{\textbf{Settings}} & \multicolumn{2}{c|}{\textbf{Training Data}}                            & \multirow{2}{*}{\textbf{Testing Data}} \\ \cline{2-3}
                                   & \multicolumn{1}{c|}{\textbf{Initial Phrase}} & \textbf{Sequent Phrase} &                                        \\ \hline
FSL                                & \multicolumn{1}{c|}{Base Classes}                    & New Classes                     & New Classes                                    \\ \hline
gFSL                               & \multicolumn{1}{c|}{Base Classes}                    & Base + New Classes               & Base + New Classes                               \\ \hline
FSCIL                              & \multicolumn{1}{c|}{Base Classes}                    & New Classes                     & Base + New Classes                               \\ \hline
\end{tabular}
\label{tab2}
\end{table}

\textbf{Few-shot Learning:} FSL is an ML problem that aims to learn a model capable of classifying and recognizing new classes with very limited training samples~\cite{chen2019closer,ye2020heterogeneous,wang2020generalizing}. Similar to FSL, FSCIL also employs $N-$way $K-$shot learning for each new class. However, FSCIL's training data comprises multiple incremental sessions, each with several few-shot classes. As Tab.~\ref{tab2} indicates, FSL's main goal is to enable model generalization to new classes using limited training data, without emphasizing base class recognition performance. In contrast, FSCIL aims to continuously learn new classes with limited samples while preserving knowledge of previously learned classes.

\textbf{General Few-shot Learning:} FSL typically doesn't consider base class performance in testing~\cite{goodfellow2013empirical}. However, real-world applications often require models to learn new classes from limited samples while maintaining performance on base classes, which often represent high-frequency classes in the real world~\cite{tao2020few,chao2016empirical}. This practical need has led to the development of a novel setting, gFSL~\cite{gidaris2018dynamic}, aimed at enabling learning of new classes with limited samples without compromising performance on previous classes~\cite{gidaris2018dynamic,qi2018low,yoon2020xtarnet}. As highlighted in Tab.~\ref{tab2}, unlike FSCIL, gFSL allows access to initial training data of base classes.

\subsection{Taxonomy}
\label{sec:over:t}
For a thorough examination of FSCIL research, we propose a taxonomy for current methods. Illustrated in Fig.~\ref{Fig7}, we analyze existing methods from three angles: data-based, structure-based, and optimization-based approaches for the FSCIC problem. Additionally, for the FSCIOD issue, we assess methods through anchor-based and anchor-free perspectives.

\begin{figure}[htbp!]
\centering
\footnotesize
\resizebox*{0.47\textwidth}{!}{\begin{tikzpicture}[xscale=0.9, yscale=0.36]

\draw [thick, -] (0, 24) -- (0, 0); \node [right] at (-0.5, 24) {\textbf{Few-shot Class-incremental Learning
}};
\draw [thick, -] (0, 23) -- (0.5, 23);\node [right] at (0.5, 23) {\textbf{Few-shot Class-incremental Classification (Section~\ref{sec:FSCIC})}};
\draw [thick, -] (1, 22.5) -- (1, 10);
\draw [thick, -] (1, 22) -- (1.5, 22);\node [right] at (1.5, 22) {Data-based Approaches (Section~\ref{sec:FSCIC:dba})};
\draw [thick, -] (2, 21.5) -- (2,18);
\draw [thick, -] (2, 21) -- (2.5, 21);\node [right] at (2.5, 21) {Data Replay-based Methods (Section~\ref{sec:FSCIC:dba:drbm})};
\draw [thick, -] (3, 20.5) -- (3,19);
\draw [thick, -] (3, 20) -- (3.5, 20);\node [right] at (3.5, 20) {Raw Replay-based Methods};
\draw [thick, -] (3, 19) -- (3.5, 19);\node [right] at (3.5, 19) {Generative Replay-based Methods
};

\draw [thick, -] (2, 18) -- (2.5, 18);\node [right] at (2.5, 18) {Pseudo Scenarios-based Methods (Section~\ref{sec:FSCIC:dba:psc})};

\draw [thick, -] (3, 17.5) -- (3, 16);
\draw [thick, -] (3, 17) -- (3.5, 17);\node [right] at (3.5, 17) {Pseudo Class-based Methods
};
\draw [thick, -] (3, 16) -- (3.5, 16);\node [right] at (3.5, 16) {Pseudo Session-based Methods
};

\draw [thick, -] (1, 15) -- (1.5, 15);\node [right] at (1.5, 15) {Structure-based Approaches (Section~\ref{sec:FSCIC:sba})};
\draw [thick, -] (2, 14.5) -- (2, 11);
\draw [thick, -] (2, 14) -- (2.5, 14);\node [right] at (2.5, 14) {Dynamic Structure-based Methods (Section~\ref{sec:FSCIC:sba:dsbm})};

\draw [thick, -] (3, 13.5) -- (3, 12);
\draw [thick, -] (3, 13) -- (3.5, 13);\node [right] at (3.5, 13) {Graph-based Methods
};
\draw [thick, -] (3, 12) -- (3.5, 12);\node [right] at (3.5, 12) {Other Methods
};

\draw [thick, -] (2, 11) -- (2.5, 11);\node [right] at (2.5, 11) {Attention-based Methods (Section~\ref{sec:FSCIC:sba:abs})};

\draw [thick, -] (1, 10) -- (1.5, 10);\node [right] at (1.5, 10) {Optimization-based Approaches (Section~\ref{sec:FSCIC:oba})};
\draw [thick, -] (2, 9.5) -- (2, 1);
\draw [thick, -] (2, 9) -- (2.5, 9);\node [right] at (2.5, 9) {Representation Learning-based Methods (Section~\ref{sec:FSCIC:oba:rlbm})};

\draw [thick, -] (3, 8.5) -- (3, 6);
\draw [thick, -] (3, 8) -- (3.5, 8);\node [right] at (3.5, 8) {Metric Learning-based Methods
};
\draw [thick, -] (3, 7) -- (3.5, 7);\node [right] at (3.5, 7) {Feature Space-based Methods
};
\draw [thick, -] (3, 6) -- (3.5, 6);\node [right] at (3.5, 6) {Feature Fusion-based Methods
};

\draw [thick, -] (2, 5) -- (2.5, 5);\node [right] at (2.5, 5) {Knowledge Distillation Methods (Section~\ref{sec:FSCIC:oba:kdbm})};

\draw [thick, -] (3, 4.5) -- (3, 3);
\draw [thick, -] (3, 4) -- (3.5, 4);\node [right] at (3.5, 4) {Knowledge Distillation-based Methods with Balanced Data
};
\draw [thick, -] (3, 3) -- (3.5, 3);\node [right] at (3.5, 3) {Optimized Knowledge Distillation-based Methods
};

\draw [thick, -] (2, 2) -- (2.5, 2);\node [right] at (2.5, 2) {Meta learning Methods (Section~\ref{sec:FSCIC:oba:mtbm})};
\draw [thick, -] (2, 1) -- (2.5, 1);\node [right] at (2.5, 1) {Other Methods (Section~\ref{sec:FSCIC:oba:om})};

\draw [thick, -] (0, 0) -- (0.5, 0);\node [right] at (0.5, 0) {\textbf{Few-shot Class-incremental Object Detection (Section~\ref{sec:FSCIOD})}};
\draw [thick, -] (1, -0.5) -- (1, -5);
\draw [thick, -] (1, -1) -- (1.5, -1);\node [right] at (1.5, -1) {Anchor-free approaches (Section~\ref{sec:FSCIOD:aff})};
\draw [thick, -] (2, -1.5) -- (2,-4);
\draw [thick, -] (2, -2) -- (2.5, -2);\node [right] at (2.5, -2) {CentreNet-based Methods
};
\draw [thick, -] (2, -3) -- (2.5, -3);\node [right] at (2.5, -3) {FCOS-based Methods
};
\draw [thick, -] (2, -4) -- (2.5, -4);\node [right] at (2.5, -4) {DETR-based Method
};
\draw [thick, -] (1, -5) -- (1.5, -5);\node [right] at (1.5, -5) {Anchor-based approaches (Section~\ref{sec:FSCIOD:abf})};
\draw [thick, -] (2, -5.5) -- (2,-6);
\draw [thick, -] (2, -6) -- (2.5, -6);\node [right] at (2.5, -6) {Mask RCNN-based Methods
};
\end{tikzpicture}}
\caption{The taxonomy of representative methods in FSCIL.}
\label{Fig7}
\end{figure}

\section{Datasets and Evaluation}
\label{sec:DE}

\subsection{Datasets}
\subsubsection{Datasets for Classification}

\textbf{\emph{mini}ImageNet:} \emph{mini}ImageNet, a diverse and challenging dataset containing object classes from various fields such as animals, plants, daily necessities, and vehicles, was originally proposed by \emph{Vinyals et al.}~\cite{vinyals2016matching} in 2016 and has been commonly used to evaluate FSL algorithms. The dataset comprises 60,000 images selected from ImageNet~\cite{russakovsky2015imagenet}, with 100 classes and 600 images per class, each sized at $84\times84$ pixels. In FSCIL, the prevalent data partitioning method by \emph{Tao et al.}~\cite{tao2020few} divides these 100 classes into 60 base and 40 incremental classes. These incremental classes are further segmented into 8 sessions, each with 5 classes and 5 training samples per class, forming a $5-$way $5-$shot setup.

\textbf{CIFAR-100:} CIFAR-100, introduced by~\emph{Krizhevsky et al.}~\cite{krizhevsky2009learning} in 2009, is widely used in CIL. It features a broad array of image data, covering classes such as plants, humans, and vehicles. The dataset comprises 100 classes, each with 600 $32\times32$ RGB images, allocated into 500 for training and 100 for testing. For FSCIL, the common data partitioning approach by \emph{Tao et al.}~\cite{tao2020few} divides these 100 classes into 60 base classes and 40 incremental classes. These incremental classes are further split into 8 sessions, each with 5 classes. Every class in these sessions has 5 training samples, establishing a $5-$way $5-$shot format.

\textbf{CUB-200:} The Caltech-UCSD Birds-200-2011 (CUB-200) dataset, created by~\emph{Wah et al.}~\cite{wah2011caltech} in 2011, is a benchmark dataset for fine-grained classification in computer vision. It comprises 11,788 images across 200 bird species. For FSCIL algorithm evaluation, the data partitioning method by~\emph{Tao et al.}~\cite{tao2020few} is commonly employed. This method splits the 200 classes into 100 base and 100 incremental classes, with these incremental classes further divided into 10 sessions. Each session encompasses 10 classes, with 10 training samples per class, resulting in each session being a $10-$way $10-$shot task. The standard image size in this context is $224\times224$ pixels.

\subsubsection{Datasets for Object Detection}
\textbf{COCO:} The Microsoft Common Objects in Context (COCO) dataset, widely used for object detection tasks, comprises 80 object classes including people, animals, vehicles, furniture, and food~\cite{lin2014microsoft}. It features a diverse and complex array of images that reflect real-world scenarios, complete with detailed annotations such as bounding boxes, class labels, and semantic segmentation masks. For FSCIOD tasks, the data partitioning strategy by \emph{Perez-Rua et al.}~\cite{perez2020incremental} is commonly used. This approach utilizes 20 classes overlapping with the PASCAL VOC dataset~\cite{everingham2009pascal} as new incremental classes and the remaining 60 as base data. FSCIOD models under this setup are evaluated using $K \in {1, 5, 10}$ bounding boxes per new class.

\textbf{PASCAL VOC:} The PASCAL Visual Object Classes (VOC) dataset, widely used for object detection tasks, includes 20 common object classes like people, animals, vehicles, and household items~\cite{everingham2009pascal}. It is frequently utilized for cross-dataset evaluations of FSCIOD algorithms. Notably, the VOC shares 20 classes with the COCO dataset. Thus, the 60 non-overlapping classes in COCO are typically the base training data for cross-dataset evaluations, with the VOC's 20 classes serving as new incremental classes to assess few-shot IL capabilities. The evaluation strategy, proposed by \emph{Perez-Rua et al.}~\cite{perez2020incremental}, is similar to that used with the COCO dataset, where FSCIOD models are evaluated using $K \in {1, 5, 10}$ bounding boxes annotated for each new class.

\subsection{Evaluation Metrics}

\subsubsection{Evaluation Metrics for Classification}
\label{emfc}
In FSCIL, the model needs to learn new classes while retaining previous knowledge. After each session, it is tested on all encountered classes, using accuracy as the primary metric. Additionally, after completing all incremental sessions, the overall performance is evaluated using Average Accuracy (AA) and Performance Dropping (PD) rate. The AA calculates the mean accuracy across the base and all incremental sessions, with higher values indicating superior performance. PD measures the accuracy drop between the base and final incremental sessions, where lower values represent better FSCIL performance. Definitions are shown in Tab.~\ref{tab33}.

\begin{table}[htbp!]
\scriptsize
\centering
\renewcommand{\arraystretch}{1}
\setlength\tabcolsep{14pt}
\caption{Evaluation metric definitions for classification and object detection tasks. $A_i$ represents the accuracy obtained in session $i$, and $B$ is the number of incremental sessions in classification. $P_i$ and $R_i$ represent the precision and recall obtained in class $i$, and $K$ is the number of counted classes.}
\begin{tabular}{|c|cl|l}
\cline{1-3}
\textbf{Task}    & \multicolumn{2}{c|}{\textbf{Metric}}                                                                                                   &  \\ \cline{1-3}
Classification   & \multicolumn{1}{c|}{\(\displaystyle AA = \frac{1}{B+1} \sum_{i=0}^{B} A_i \)} & \(\displaystyle PD = A_B - A_0 \)                      &  \\ \cline{1-3}
Object Detection & \multicolumn{1}{c|}{\(\displaystyle mAP = \frac{1}{K} \sum_{i=1}^{K} P_i \)}   & \(\displaystyle mAR = \frac{1}{K} \sum_{i=1}^{K} R_i \) &  \\ \cline{1-3}
\end{tabular}
\label{tab33}
\end{table}

\subsubsection{Evaluation Metrics for Object Detection}
In FSCIOD tasks, two approaches incorporate new incremental data: batch and continuous IL. Batch IL entails learning all new classes at once, while continuous IL adds new classes progressively. Batch IL, similar to single-session FSCIL, is more common. The predominant performance metric is mean Average Precision (mAP), which is the mean of the AP values calculated for all counted classes. mAP is calculated separately for base classes, new classes, and all classes, with higher mAP values across all classes indicating better FSCIOD performance. Additionally, some studies use a similar way to calculate mean Average Recall (mAR) and mAP50 as complementary metrics for a more comprehensive evaluation. Definitions are shown in Tab.~\ref{tab33}.

\subsection{Summary}
The overview of datasets and evaluation methods reveals a scarcity of publicly available datasets for FSCIL tasks, limiting their practical application. Some studies, such as~\cite{pan2023ssfe}, have introduced datasets for various FSCIL scenarios, but there remains significant scope for dataset enhancement. Regarding model evaluation, while current metrics assess the model's learning ability to some extent, they don't completely capture the detailed performance of FSCIL throughout the continuous learning process~\cite{peng2022few}. Hence, both datasets and evaluation metrics in FSCIL present substantial opportunities for further development.

\section{Few-shot Class-incremental Classification}
\label{sec:FSCIC}

This section, focusing on FSCIL classification tasks, summarizes existing methods classified into data-based, structure-based, and optimization-based categories, noting some overlap across these domains. The methods are categorized based on their attributes and core innovations, concluding with a performance comparison and key concerns discussion.

\begin{figure}[htbp!]
\centering
\includegraphics[width=0.45\textwidth]{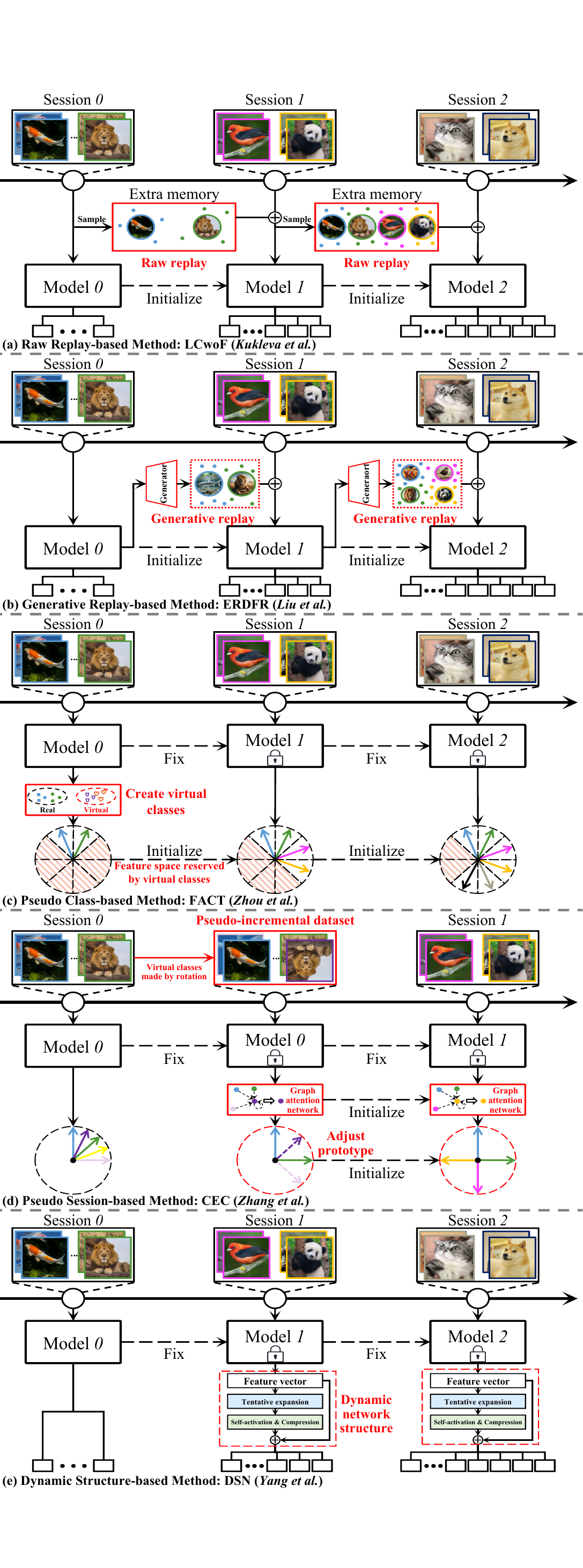}
\caption{The representative classification methods in FSCIL, with the core designs highlighted in red. (a) LCwoF stores some old samples for raw replay, jointly calibrating performance with the new classes during new sessions; (b) ERDFR trains a generator using the previous main model and synthesizes virtual old samples for DR during the learning of new classes, aiding in the retention of old knowledge; (c) FACT creates virtual incremental classes from base classes to simulate future scenarios, enabling the model to develop forward compatibility during the base session; (d) CEC constructs a pseudo-incremental session from the base session to train a GAT, which is later used to adjust the relationship between new and old class prototypes during real incremental sessions; (e) DSN designs a dynamic structure alongside the backbone that allows for expansion and autonomous compression, facilitating the learning of new classes while retaining old knowledge.}
\label{Fig8}
\end{figure}

\subsection{Data-based Approaches}
\label{sec:FSCIC:dba}
Data-based approaches refer to addressing FSCIL challenges arising from limited or non-reusable data by focusing on the data perspective. Relevant methodologies include DR and pseudo-data construction.

\subsubsection{Data Replay-based Methods}
\label{sec:FSCIC:dba:drbm}
Catastrophic forgetting often occurs in FSCIL due to the unavailability of original complete training data from previous sessions. DR is a direct strategy to mitigate this issue by replaying valuable data while adapting to new sessions. Existing methods include raw replay and generative replay, involving the replay of samples or feature representations.

\textbf{Raw Replay-based Methods:}
The raw replay methods address catastrophic forgetting by storing a portion of raw samples from previous sessions in auxiliary memory and replaying it during the learning process of a new session to review previous knowledge. As shown in Fig.~\ref{Fig8}(a), \emph{Kukleva et al.}~\cite{kukleva2021generalized} proposed a multi-stage FSCIL method called LCwoF. It first used the Cross-entropy (CE) loss to train the backbone. In the second stage, it employed the KD loss and base-normalized CE loss to jointly supervise learning new classes and preserve the old knowledge. In the final stage, randomly sampled old and new class data were combined for DR to further calibrate the performance. Differently, \emph{Zhu et al.}~\cite{zhu2022feature} proposed a feature distribution distillation-based method, which stored the same number of old samples as each new class to form a joint set during its learning process. Both the old and new models generated the feature representations for this set. A joint function based on CE loss and KD loss was employed to constrain the new model to generate similar representations as the old model to preserve the old knowledge. However, the performance of raw replay methods is influenced by factors such as auxiliary storage space, sample selection, and quantity, which have not been fully addressed.

\textbf{Generative Replay-based Methods:}
The generative replay methods train and store a model that generates data, including samples or feature representations of old classes during the new session learning process to review old knowledge. As shown in Fig.~\ref{Fig8}(b), \emph{Liu et al.}~\cite{liu2022few} proposed a data-free replay method that used GAN-like ideas to train a generator with an uncertainty constraint based on entropy regularization so that the generated data could get close to the decision boundary. In incremental sessions, the generated and new data fine-tuned the model, giving it a good performance on new and old classes. Different from generating samples, some methods choose to generate features. Specifically, \emph{Shankarampeta and Yamauchi}~\cite{shankarampeta2021few} proposed a framework based on Wasserstein GAN~\cite{arjovsky2017wasserstein} with MAML~\cite{finn2017model}, which mainly consisted of a feature extractor and a feature generator. During the training process, the feature extractor was initialized on base data, and then the feature generator was trained by meta-learning with MAML. In IL, the feature extractor with feature distillation was combined with feature replay at the classifier level to tackle catastrophic forgetting. Similarly, FSIL-GAN proposed by \emph{Agarwal et al.}~\cite{agarwal2022semantics} used a similar framework to perform feature replay. The main contribution of FSIL-GAN was the import of the semantic projection module, which constrained the synthesized features to match with the latent semantic vectors to ensure their diversity and discriminability. In IL, KD ensured knowledge transfer between the old and new generators. Generative replay offers flexibility and safety in sample generation but increases the model complexity.

\textbf{Discussion:}
DR is a direct strategy to address catastrophic forgetting in FSCIL. While raw replay methods provide simplicity and convenience, their effectiveness is influenced by factors including auxiliary storage space, the selection and quantity of samples, and the imbalanced distribution of old and new classes. In comparison, generative replay methods exhibit better flexibility and mitigate potential privacy concerns associated with raw replay methods. However, generative replay methods face challenges in continuously generating old samples in the imbalanced and dynamic data stream, generation quality and efficiency, and additional computational costs. These issues require further exploration and research.

\subsubsection{Pseudo Scenarios-based Methods}
\label{sec:FSCIC:dba:psc}

Contrasting backward-compatible methods such as DR that tackle catastrophic forgetting, another prevalent FSCIL strategy is the construction of pseudo-incremental scenarios. These scenarios, acting as preview mechanisms in the dynamic and ever-expanding FSCIL data stream, prepare models for actual incremental sessions, ensuring effective performance. These methods primarily fall into two categories: pseudo-class and pseudo-session construction.

\textbf{Pseudo Class-based Methods:}
Pseudo-class construction methods aim to generate synthetic classes and their corresponding samples to facilitate FSCIL models preparing for the real incremental classes. Most current studies employ base sessions to develop these pseudo-classes, training the models using pseudo-data and the original data. This strategy promotes forward compatibility in the FSCIL models. As shown in Fig.~\ref{Fig8}(c), this approach is the forward-compatible FSCIL framework proposed by \emph{Zhou et al.}~\cite{zhou2022forward}. The crux of this framework lay in constraining the real samples during training, enabling them to render their respective categories more compact in the embedding space and reserve some spaces for the constructed virtual categories. In particular, this method promoted the intra-class compactness of the real data and forced the masked features based on the real data to be closed to a pseudo-class. Simultaneously, the framework employed similar techniques to constrain virtual features constructed from a mixture of multiple class features. It ensured the compactness of real categories while reserving some feature space for incremental classes. Similarly, \emph{Peng et al.}~\cite{peng2022few} generated pseudo-classes by merging two distinct classes from the base session and augmenting the data using techniques such as random cropping, horizontal flipping, and color jittering in the ALICE framework. It used angular penalty loss commonly used in face recognition for feature extractor training based on the joint set of pseudo and real data. The core idea also involved promoting intra-class compactness and reserving feature space for incremental classes.

\textbf{Pseudo Session-based Methods:}
Unlike pseudo-class methods that create synthetic classes, pseudo-session construction methods focus more on emulating incremental sessions. Most existing approaches use base sessions to create pseudo-incremental sessions and meta-learning techniques to allow FSCIL models to understand how to handle incremental sessions. The ways to construct pseudo-incremental sessions are various. As shown in Fig.~\ref{Fig8}(d), the CEC framework proposed by \emph{Zhang et al.}~\cite{zhang2021few} applied the large angle rotation transformation on the base classes to build pseudo-incremental sessions. These pseudo-sessions were then combined with the base session to train the graph attention network by meta-learning strategy so that it could pass context information between prototypes, thus better equipping it to handle the FSCIL task. The FSCIL model by \emph{Zhu et al.}~\cite{zhu2021self} included two innovations: Random Episode Selection (RES) and Dynamic Relation Projection (DRP). RES sampled five classes randomly to create $N-$way $K-$shot pseudo-incremental sessions, masking original class prototypes and using pseudo-incremental data to generate class prototypes by averaging. These prototypes were refined using DRP, which mapped class prototypes from standard and pseudo-IL to shared latent space. It calculated the cosine similarity between old and new classes to obtain a relation matrix. This matrix acted as a transitional coefficient for prototype optimization, enabling dynamic optimization to preserve existing knowledge and boost new classes' discriminative ability.

\textbf{Discussion:}
Pseudo-scenario construction is a forward-compatible strategy, synthesizing classes or sessions to train models for future real incremental classes. Pseudo-class construction is a method where pseudo-classes are constructed in conjunction with base classes to train the model, enabling the feature space to reserve certain spaces for upcoming incremental classes. However, reserving space often requires prior knowledge of the total number of incremental classes, which contradicts the real world. Since synthetic and real data often exhibit differences, the suitability of reserved space remains to be discovered. In contrast, pseudo-session construction is more reasonable, as it often combines the pseudo-incremental sessions with meta-learning to train the model that can learn to adapt to incremental sessions. However, the issue of whether pseudo-incremental sessions can effectively simulate real incremental sessions needs further exploration.

\subsection{Structure-based Approaches}
\label{sec:FSCIC:sba}
Structure-based approaches refer to utilizing the model design or its characteristics to address the challenges in FSCIL. These methods mainly involve dynamic structure methods and attention-based methods.

\subsubsection{Dynamic Structure-based Methods}
\label{sec:FSCIC:sba:dsbm}
Dynamic structure methods aim to achieve FSCIL by dynamically adjusting the model structure or the interrelationships between prototypes. Currently, existing methods can be broadly categorized into graph-based and other methods.

\textbf{Graph-based Methods:}
Methods based on graph structures utilize graph topological properties to achieve FSCIL. These methods typically use nodes and edges in the graph to describe the similarity or correlation between different classes from various sessions and adjust the graph structure based on the mutual influences among classes. Some studies employ graph structures to implement FSCIL~\cite{tao2020few,zhang2021few}. For example, \emph{Tao et al.}~\cite{tao2020few} proposed the TOPIC framework, which utilized the neural gas network for knowledge extraction and representation. TOPIC aimed to address FSCIL by dynamically adjusting the interrelationships between feature representations. Specifically, the neural gas network defined an undirected graph that mapped the feature space to a finite set of feature vectors and maintained the topological properties of the feature space through competitive Hebbian learning~\cite{martinetz1993competitive}. To achieve FSCIL, they gradually improved the neural gas network by enabling the supervised neural gas model to grow nodes and edges through competitive learning. Additionally, they designed a stability loss to suppress catastrophic forgetting and an adaptability loss to reduce overfitting. In addition, the CEC framework~\cite{zhang2021few} mentioned in Sec.~\ref{sec:FSCIC:dba:psc} also utilized graph structures for FSCIL. It first trained a graph attention network using pseudo-incremental sessions to adjust the model. During incremental sessions, the model incorporated an attention mechanism to regulate the interrelationships between nodes, represented by prototypes of old and new classes. This allowed for better context information transfer between sessions, making the class prototypes more robust.

\textbf{Other Methods:}
In addition to graph-based methods, some studies employ other dynamic structures to achieve FSCIL~\cite{yang2021learnable,yang2022dynamic,ahmad2022few}. For example, \emph{Yang et al.} proposed a series of works~\cite{yang2021learnable,yang2022dynamic}. As shown in Fig.~\ref{Fig8}(e), they proposed a novel Dynamic Support Network (DSN)~\cite{yang2022dynamic} to address the challenges of FSCIL. DSN was a self-adaptive updating network with compressed node expansion, aiming to "support" the feature space. In each session, DSN temporarily expanded the network nodes to enhance the feature representation capability for incremental classes. Then, it dynamically compressed the expanded network through node self-activation, pursuing compact feature representation to alleviate overfitting. Moreover, DSN selectively invoked the distribution of old classes during the IL process to support feature distribution and avoid confusion between classes. Furthermore, in the framework proposed by \emph{Ahmad et al.}~\cite{ahmad2022few}, the output nodes of the model increased with the number of classes involved in the current session. The model parameters related to old classes were kept fixed. The newly added nodes' weights were randomly initialized, and they were trained using the training data from the current session to update the parameters.

\textbf{Discussion:}
Dynamic structure methods are important approaches to address the challenges of FSCIL. These methods achieve the learning of new knowledge while preserving old knowledge by dynamically adjusting the model structure or the relationships between prototypes. For example, graph-based methods utilize the topological characteristics of graphs to achieve non-forgettable IL by adjusting nodes and edges to describe the similarity and correlation between different classes. Dynamic structure networks enhances feature representation and alleviates overfitting by temporarily expanding and dynamically compressing network nodes. Dynamic structural methods play a significant role in FSCIL, but further research and exploration are still needed to develop more design methods for dynamic structures.

\subsubsection{Attention-based Methods}

\label{sec:FSCIC:sba:abs}

Attention-based methods in FSCIL adjust the attention allocation of features by introducing attention modules into the model structure. This allows the model to focus on information relevant to the current task, improving its performance and generalization ability. The role of attention modules used in many FSCIL approaches~\cite{zhao2023few,zhang2021few,zhou2022few,chi2022metafscil,cheraghian2021semantic} is diverse. For example, in the dual-branch KD framework proposed by \emph{Zhao et al.}~\cite{zhao2023few}, which consisted of a base branch and a novel branch, they noted that fine-tuning by novel classes inevitably led to forgetting old knowledge. To further improve the performance of base classes, they proposed an attention-based aggregation module that selectively merges predictions from the base branch and the novel branch. Furthermore, \emph{Cheraghian et al.}~\cite{cheraghian2021semantic} employed meta-learning to train a backbone that can incrementally learn new classes with limited samples without forgetting the old classes. However, many existing FSCIL paradigms updated the classifier by concatenating the base classifier with the new class prototypes obtained by averaging the features of each training sample. This approach often led to bias. Therefore, this paper proposed a correction model based on Transformer~\cite{vaswani2017attention}. With its attention mechanism, the correction model can effectively transmit context information among different classes, making the classifier more efficient and robust. Similar approaches included the graph attention network used in the CEC framework~\cite{zhang2021few}.

\subsection{Optimization-based Approaches}
\label{sec:FSCIC:oba}
Optimization-based approaches tackle the challenges in FSCIL by addressing the complexity of optimization problems. The relevant strategies primarily involve representation learning, meta-learning, and KD, according to the existing works.

\subsubsection{Representation Learning-based Methods}
\label{sec:FSCIC:oba:rlbm}
In FSCIL, representation learning aims to extract meaningful features from a limited stream of samples to form a "representation" of the data~\cite{bengio2013representation}. Through effective representation learning, models can identify and utilize underlying patterns within these few samples and generalize them to new, unseen classes. Even in few-shot incremental scenarios, models can perform excellently with efficient representation learning. In FSCIL, there are diverse approaches to performing representation learning, which can be categorized into metric learning-based, feature space-based, feature fusion-based, and other methods, based on the core principles of the respective methods.

\begin{figure}[h!]
\centering
\includegraphics[width=0.45\textwidth]{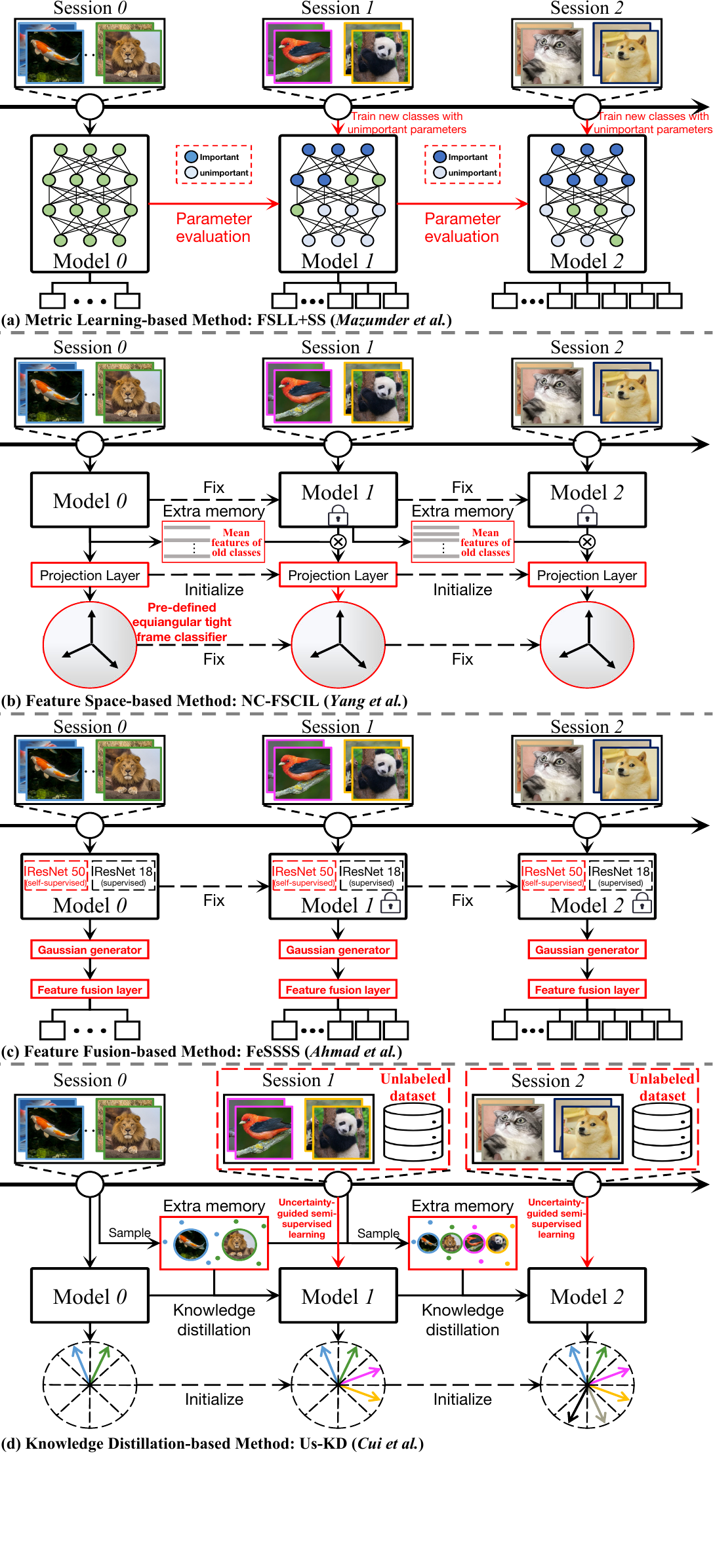}

\caption{The representative classification methods in FSCIL, with the core designs highlighted in red. (a) FSLL+SS initially uses self-supervised learning to provide good generalization. Then, it evaluates the importance of model parameters, fixing important ones to maintain old knowledge while using metric learning to learn new classes with unimportant parameters; (b) NC-FSCIL uses neural collapse theory to pre-define a classifier before training, and the model is constrained to learn towards the pre-defined classifier via the projection layer, preventing conflicts between learning new and old classes; (c) FeSSSS introduces a ResNet50 trained through self-supervised learning in addition to the regular ResNet backbone. By merging the supervised and self-supervised features, the framework enhances the generalization capability in FSCIL; (d) Us-KD focuses on semi-supervised FSCIL, first utilizing KD with stored old samples and new data to learn new knowledge while retaining old knowledge. Then, uncertainty quantification is applied to select suitable unlabeled data for labeling, combined with labeled samples for iterative model updates.}
\label{Fig9}
\end{figure}

\textbf{Metric Learning-based Methods:}
Metric learning aims to determine the similarity between objects using an optimal distance metric for learning tasks~\cite{kaya2019deep}. It has found extensive application in FSL~\cite{li2023deep}. Recently, metric learning has also been adopted in FSCIL to learn effective representations. Among the commonly used approaches, triplet loss stands out. As shown in Fig.~\ref{Fig9}(a), \emph{Mazumder et al.}~\cite{mazumder2021few} proposed a novel approach for FSCIL. It incorporated self-supervised learning to enhance the generalization capability of the backbone. Then, this approach analyzed the importance of model parameters and updated only the unimportant parameters for new classes. The update was achieved by combining three loss functions: triplet loss, regularization loss, and cosine similarity loss. The triplet loss aimed to generate discriminative features, while the regularization loss prevented catastrophic forgetting. The cosine similarity loss focused on controlling the similarity between old and new prototypes. Thus, FSCIL achieved effective performance. Moreover, other metric learning methods are applied to FSCIL too. Concretely, \emph{Peng et al.}~\cite{peng2022few} proposed the ALICE framework, incorporating the angular penalty loss originally used in face recognition to obtain well-clustered features. This loss was employed to train the backbone using both base class data and synthetic data, thereby creating additional space for accommodating incremental classes, and cosine similarity was applied to achieve the classification. 

Despite the good performance achieved by these margin-based metric losses, \emph{Zou et al.}~\cite{zou2022margin} pointed out the issue in FSCIL: large margin values can result in good discriminability among base classes but hinder the generalization capability of new classes. Conversely, small or even negative margin values can lead to poor performance on the base classes but exhibit better generalization on new classes. This phenomenon is known as the class-level overfitting problem. To address this issue, \emph{Zou et al.}~\cite{zou2022margin} proposed the CLOM framework, which combined margin theory with the characteristics of neural network structures. Specifically, since the shallow layers of neural networks are more suitable for learning common features among classes, while the deep layers are better suited to acquiring advanced features, they designed a loss function that constrains shallow feature learning and deep feature learning separately. Furthermore, this framework alleviated the class-level overfitting problem by integrating class relationships.

\textbf{Feature Space-based Methods:}
Feature space-based methods are a class of approaches that aim to perform FSCIL by optimizing the feature space. The core idea of these methods is to design the feature space for learning more robust and efficient feature representations. Some related methods address FSCIL by designing subspaces~\cite{zhao2021mgsvf,cheraghian2021synthesized,kim2023warping}. For example, inspired by the frequency decoupling~\cite{khorramshahi2020devil}, \emph{Zhao et al.}~\cite{zhao2021mgsvf} discussed and utilized the characteristics of different frequency components in features. Specifically, the method first trained a feature extractor using metric learning loss and regularization loss. Then, they decoupled the features based on their frequency and observed the roles of high-frequency and low-frequency information in FSCIL. It was found that low-frequency information may contribute more to preserving old knowledge. Therefore, they designed subspaces with different learning rates to learn features in different frequency domains, where the fast subspace learned new knowledge and the slow subspace preserved old knowledge. Through this subspace combination strategy, the method achieved good performance.

Furthermore, some methods address FSCIL by designing feature spaces of specialized structures. For example, \emph{Hersche et al.}~\cite{hersche2022constrained} proposed a C-FSCIL framework, which consisted of a feature extractor trained by meta-learning, a trainable fully connected layer, and a rewritable explicit memory. The core idea was introducing hyperdimensional embedding, which has three advantages: (1) the high probability of quasi-orthogonality between random vectors, (2) rich expressive space, and (3) good semantic representation capability. C-FSCIL had three training strategies. The first was based on simple meta-learning, as described in Sec.~\ref{sec:FSCIC:oba:mtbm}. The second strategy stored initial prototypes in the globally average activation memory and applied an element-wise sign operation to transform similar feature prototypes into bipolar vectors. These transformed vectors were then supervised to train the fully connected layer, which learned the weights for the final prototypes. The third strategy was similar to the second one but incorporated two losses to constrain the inter-class differences and maintain the relationship with the original prototypes. Besides, as shown in Fig.~\ref{Fig9}(b), \emph{Yang et al.}~\cite{yang2023neural} proposed an FSCIL framework based on neural collapse~\cite{papyan2020prevalence}, which refers to the phenomenon that at the end of training (after 0 training error rate), the last layer features of the same class collapse into a single vertex in the feature space, aligning all class vertices with their classifier prototypes and forming as a simplex equiangular tight frame. Based on this characteristic, the proposed framework predefined a structure similar to neural collapse and directed the model to optimize it. Specifically, a group of prototypes for both the base and incremental sessions was pre-assigned as a simplified form of equiangular tight frame. During training, the prototypes were fixed. They introduced a novel loss function and an additional projection layer to assign each class to its respective prototype separately. Without cumbersome operations, this method achieved superior performance. In addition, in Sec.~\ref{sec:FSCIC:dba:psc}, the method proposed by \emph{Zhou et al.}~\cite{zhou2022forward} also addressed FSCIL by learning the feature space in a way that preserves some space for incremental classes during the learning of base classes.

\textbf{Feature Fusion-based Methods:}
Feature fusion refers to integrating or combining features obtained from different information sources or feature extraction methods to create a more comprehensive and efficient representation that exhibits robustness and generalization capabilities~\cite{sankaran2021feature}. In the context of FSCIL, various methods employ feature fusion strategies to learn effective feature representations that can adapt to specific task requirements. Notably, a significant focus is on incorporating self-supervised features into the fusion process~\cite{mazumder2021few,ahmad2022few,ahmad2022variable,kalla2022s3c}. For example, as shown in Fig.~\ref{Fig9}(c), \emph{Ahmad et al.}~\cite{ahmad2022few} proposed a framework that combines self-supervised and supervised features. The core structure of this framework included the following components: Firstly, feature extractors obtained through supervised training on base-class data and self-supervised tasks on ImageNet~\cite{russakovsky2015imagenet} or OpenImages-v6~\cite{kuznetsova2020open} using methods such as pretext tasks, contrastive loss, or clustering. Secondly, the Gaussian generator synthesized feature for replay in incremental sessions. Lastly, a lightweight model for incremental feature fusion and classification. Additionally, \emph{Kalla and Biswas}~\cite{kalla2022s3c} proposed S3C, a method for addressing FSCIL based on the stochastic classifier~\cite{neal2012bayesian} and self-supervision. They introduced a novel self-supervised training approach~\cite{lee2020self}, using image augmentations to generate artificial labels, to train the classification layer. The stochastic classifier weights helped mitigate the impact of limited new samples and the unavailability of old samples. The self-supervision component enabled the learning of features from base classes that generalize well to future unseen classes, effectively reducing catastrophic forgetting.

In addition to feature fusion based on self-supervised features, there are also works that integrate other features to achieve good performance in FSCIL. For example, \emph{Yao et al.}~\cite{yao2022few} proposed a simple strategy for enhancing the new prototype. Specifically, it first trained a Convolutional Neural Network (CNN) on the base classes and kept it fixed. It was used to generate class prototypes for both base and new classes. Then, the initial class prototypes for new classes were measured for their similarity to base class prototypes. Based on the similarity, the prototypes for new classes were updated by mixing the initial weights with other base prototypes. This fusion enhancement strategy imitated human cognition by guiding new class learning based on existing knowledge.

\textbf{Discussion:}
Feature fusion plays a crucial role in FSCIL by integrating multiple information sources and feature extraction methods to provide a comprehensive, efficient, and robust representation. In FSCIL, various feature fusion strategies are employed to learn effective feature representations that adapt to specific tasks. For instance, combining self-supervised and supervised features enables the model to acquire representations with good generalization ability. Additionally, another approach fuses existing features to guide new class learning. These methods highlight the significance of feature fusion in addressing FSCIL challenges, while further exploration of more efficient feature fusion strategies is needed to enhance model performance and generalization ability.

\subsubsection{Knowledge Distillation-based Methods}
\label{sec:FSCIC:oba:kdbm}
In continuous learning, KD is widely employed to transfer knowledge from an old model, known as the ``teacher model," to a new model, referred to as the ``student model"~\cite{gou2021knowledge}. It effectively addresses catastrophic forgetting in continuous learning. However, in FSCIL, the data distribution between the base and incremental sessions is imbalanced, with sufficient samples in the base session and limited samples in the incremental session. Conventional KD methods for continuous learning are prone to overfitting in the incremental session and further biases in future incremental sessions~\cite{zhao2023few}. Nevertheless, many studies have explored the application of KD to FSCIL, focusing on transferring knowledge between sessions using KD. Based on the approach to address the challenges of data imbalance and overfitting in FSCIL, these studies can be classified into two categories: KD by balancing data and optimized KD.

\textbf{Knowledge Distillation-based Methods with Balanced Data:} To address the inadequacy of KD methods in FSCIL due to data imbalance, some approaches~\cite{kukleva2021generalized,dong2021few,zhu2022feature} address the issue by selecting an equal number of samples from the base and incremental sessions for distillation, avoiding bias. For instance, \emph{Dong et al.}~\cite{dong2021few} proposed a relation KD framework. They constructed a sample relation graph to represent learned knowledge, ensuring balance by selecting an equal number of samples from each base class. The samples were chosen based on the angles between their feature vectors, removing redundancies until the desired $K$ samples remained. A sample relation loss function was introduced to discover the relationship knowledge among different classes, facilitating the distillation of sample relationships and the propagation of structural information in the graph. Additionally, as introduced in Sec.~\ref{sec:FSCIC:dba:drbm}, \emph{Zhu et al.}~\cite{zhu2022feature} addressed overfitting and knowledge transfer in FSCIL by fine-tuning the backbone and sampling base class samples.

Another solution to mitigate data imbalance and limited samples in FSCIL is introducing additional data to prevent overfitting. In the context of FSCIL, \emph{Cui et al.} proposed a series of semi-supervised methods that leverage KD and unlabeled data~\cite{cui2021semi,cui2022uncertainty,cui2023uncertainty}. As shown in Fig.~\ref{Fig9}(d), \emph{Cui et al.}~\cite{cui2022uncertainty} introduced the Us-KD framework, which used an uncertainty-guided module to select unlabeled data to mitigate overfitting during knowledge transfer. The framework initially trained the model on base classes and stored some samples. In the incremental session, the model was initialized with the previous model's weights and updated using stored old samples and labeled samples from the current session through classification and distillation losses. Subsequently, the uncertainty-guided module selected and labeled unlabeled samples, combined with labeled samples to update the model iteratively. Finally, the stored data was updated with these samples. In their further research~\cite{cui2023uncertainty}, they pointed out that well-learned or easily classifiable classes often have higher prediction probabilities. Thus, they designed a data selection method called "Class Equilibrium," where well-learned categories were assigned fewer samples, and poorly learned categories were assigned more samples. It is worth noting that they highlighted the potential unreliability of KD with unlabeled data. Thus, they introduced an uncertainty-aware distillation approach suitable for semi-supervised FSCIL, consisting of uncertainty-guided refinement and adaptive distillation loss. Refinement involved leveraging uncertainty assessment to filter reliable samples from the augmented dataset, while adaptive distillation adjusted the distillation loss weights based on the sample quantity.

\textbf{Optimized Knowledge Distillation-based Methods:} Some methods have innovatively optimized KD methods to address the issues caused by the characteristics of the FSCIL data stream. For example, \emph{Cheraghian et al.}~\cite{cheraghian2021semantic} proposed a semantic-aware KD framework. In the base session training, this framework first mapped the labels to word embeddings using natural language processing models. Then, the backbone was used to convert images into original features. Subsequently, a multi-head attention model was trained using a super-class aggregation approach to prevent overfitting during the incremental process. Finally, a mapping model was trained to align the image features with the word embeddings. For incremental sessions, the labels of the new classes were first mapped to word embeddings. The mapping model was trained using fine-tuning and KD to further refine the image features. The classification was achieved by assessing the similarity between the image features and word embeddings. Additionally, \emph{Zhao et al.}~\cite{zhao2023few} proposed a class-aware bilateral distillation framework, which consists of two branches: the base branch and the novel branch. Two teacher models guided the learning of the novel branch. One teacher model was trained on base class data and possessed rich general knowledge to alleviate the overfitting of new classes. The other teacher model was updated from the previous incremental session and contained adaptive knowledge of the previous new classes to mitigate their forgetting. Fine-tuning leads to forgetting, and an attention-based aggregation module was inevitably introduced to further improve the performance of the base classes by selectively merging the predictions from the base branch and the novel branch.

\textbf{Discussion:} The applicability of KD in FSCIL depends on resolving challenges of imbalanced data distribution and overfitting with few-shot samples and its exacerbated effects resulting from incremental scenarios. The data-driven approach can address these challenges, including incorporating unlabeled data, establishing a balanced distribution, and employing sample relation distillation. Furthermore, optimizing the KD framework is another strategy. For instance, introducing semantic word embeddings as auxiliary information can be employed to optimize. These approaches aim to alleviate the above challenges and facilitate the effective application of KD in FSCIL.

\subsubsection{Meta Learning-based Methods}
\label{sec:FSCIC:oba:mtbm}
FSCIL faces challenges of overfitting and catastrophic forgetting due to limited samples for continuous learning. Meta-learning, or "learning to learn," is a prominent approach to address these issues. Meta-learning leverages experiences distilled from multiple learning episodes, encompassing a distribution of related tasks, to enhance future learning performance~\cite{hospedales2021meta}. In FSCIL, meta-learning is crucial in improving the model's adaptation ability. Building on the description in Sec.~\ref{sec:FSCIC:dba:psc}, most meta-learning methods in FSCIL are trained by pseudo-incremental tasks sampled from the base session. It proves effective for backbone training, special structure training, feature distribution learning, and various other applications in FSCIL.

One common application of meta-learning is to directly train backbone models by constructing a series of pseudo-incremental scenarios, enabling them to adapt to real incremental scenarios. For instance, in the C-FSCIL framework proposed by \emph{Hersche et al.}~\cite{hersche2022constrained}, empirical evidence demonstrated that training the backbone using the meta-learning strategy effectively can extract robust features. Utilizing the average of these features to create class prototypes surpassed the state-of-the-art methods at that time. Moreover, meta-learning was employed to learn feature distributions in FSCIL. \emph{Zheng and Zhang}~\cite{zheng2021few} introduced meta-learned class structures to regulate the distribution of learned classes in the feature space. Class structures describe the distribution of learned classes in specific directions. They ensured discriminative class prototypes without interference by proposing a class structure regularizer consisting of direction vectors associated with class structures and an alignment kernel aligning sampled embeddings with the class structures. A novel loss function was also introduced to prevent interference between new and old prototypes. The model was trained on a series of constructed meta-learning tasks. Additionally, meta-learning can be utilized to train specially designed structures in FSCIL. In the LIMIT framework proposed by \emph{Zhou et al.}~\cite{zhou2022few}, a series of pseudo-incremental tasks were sampled from the base session for meta-learning-based training. To mitigate bias issues caused by direct classification, a corrective model with a transformer as its core was introduced. The corrective model, trained using meta-learning and incorporating self-attention mechanisms, adjusted the biased relationship between old class classifiers and new class prototypes, ensuring that feature embeddings encompass contextual information. Similarly, the CEC framework mentioned in Sec.~\ref{sec:FSCIC:dba:psc} combined pseudo-incremental sessions with meta-learning to train a graph attention network for regulating the relationships between prototypes.

\subsubsection{Other Methods}
\label{sec:FSCIC:oba:om}
In addition to the methods above, some studies focus on learning efficient feature representations to adapt to FSCIL through other approaches. For instance, unlike existing methods that attempt to overcome catastrophic forgetting when learning new tasks, \emph{Shi et al.}~\cite{shi2021overcoming} proposed a novel strategy to address this issue while learning base classes. The core idea was to identify the flat local minima of the loss function during base training and perform fine-tuning in the flat region during incremental sessions. This approach maximized the preservation of knowledge when conducting fine-tuning on novel classes. Specifically, since directly finding the flat local minima is challenging, they proposed adding random noise to the model parameters to approximate it during base training. In the incremental sessions, FSCIL was achieved through fine-tuning within the flat local range. The experiments showed effectiveness.

\subsection{Summary}
\subsubsection{Performance Comparison}
In this section, we summarize the performance of mainstream FSCIC methods. Since not all relevant methods are open-source and their implementation conditions and configurations (such as different backbone networks, feature fusion with other methods, and different learning paradigms) vary, we summarize the performance of mainstream FSCIC methods with similar backbones on three commonly used benchmark datasets in Table~\ref{tab3}, including \emph{mini}ImageNet, CIFAR-100, and CUB-200, to enable a fair comparison as much as possible. The performance values are extracted from corresponding papers. To fully demonstrate the characteristics of each method, Tab.~\ref{tab3} provides their types and specific taxonomy categories. In addition, we provide the backbone used by each method in this table. For methods with too many additional factors, we provide a supplementary table in the appendix for reference. Since some methods introduce extra auxiliary factors, we have specially included an "extra factor" column in the table to summarize these factors for each method. The performance of FSCIC methods is primarily evaluated by measuring the accuracy achieved on different incremental sessions, AA across all sessions, and PD values. Given the space limitations, we only provide accuracy for the starting and ending sessions, AA, and PD. Moreover, we summarize the highlights of each method in these tables.

\begin{table*}[htp]
\fontsize{6}{6}\selectfont
\centering
\renewcommand{\arraystretch}{1.2}
\setlength\tabcolsep{0.8pt}
\caption{The performance of mainstream FSCIC methods. The data are extracted from the original papers. For \emph{mini}ImageNet and CUB-200, all the methods employ ResNet18 as the backbone. For CIFAR-100, related methods have two different settings, ResNet18 and ResNet20. To provide a comprehensive comparison, we report them in this table together, with the details noted. The accuracy of the first and last sessions is abbreviated as SA and EA. We use “-” to mark the dataset without reporting in the original papers. The best results are bold and underlined, while the second-best are underlined only. (In \%)}
\begin{tabular}{|c|l|l|l|ccccc|ccccc|ccccc|l|}
\hline
\multirow{2}{*}{\textbf{Type}}       & \multirow{2}{*}{\textbf{Taxonomy}}                                                       & \multirow{2}{*}{\textbf{Methods}} & \multirow{2}{*}{\textbf{Venue}} & \multicolumn{5}{c|}{\textbf{\emph{mini}ImageNet}}                                                                                                                                & \multicolumn{5}{c|}{\textbf{CIFAR-100}}                                                                                                                                            & \multicolumn{5}{c|}{\textbf{CUB-200}}                                                                                                                                     & \multicolumn{1}{c|}{\multirow{2}{*}{\textbf{Highlights}}}                                                                                                                                         \\ \cline{5-19}
                                     &                                                                                          &                                   &                                 & \multicolumn{1}{c|}{\scalebox{0.6}{\textbf{ \begin{tabular}[c]{@{}c@{}}Backbone\\(ResNet)\end{tabular}}}} & \multicolumn{1}{c|}{\textbf{SA}}    & \multicolumn{1}{c|}{\textbf{EA}}    & \multicolumn{1}{c|}{\textbf{AA}}    & \textbf{PD↓}   & \multicolumn{1}{c|}{\scalebox{0.6}{\textbf{ \begin{tabular}[c]{@{}c@{}}Backbone\\(ResNet)\end{tabular}}}} & \multicolumn{1}{c|}{\textbf{SA}}    & \multicolumn{1}{c|}{\textbf{EA}}    & \multicolumn{1}{c|}{\textbf{AA}} & \textbf{PD↓}               & \multicolumn{1}{c|}{\scalebox{0.6}{\textbf{ \begin{tabular}[c]{@{}c@{}}Backbone\\(ResNet)\end{tabular}}}} & \multicolumn{1}{c|}{\textbf{SA}}    & \multicolumn{1}{c|}{\textbf{EA}}    & \multicolumn{1}{c|}{\textbf{AA}}    & \textbf{PD↓}   & \multicolumn{1}{c|}{}                                                                                                                                                                             \\ \hline
\multirow{7}{*}{\rotatebox{90}{Data-based}} & Data Replay                                                                              & FSIL-GAN~\cite{agarwal2022semantics}                           & \scalebox{0.8}{ACM MM 22}                          & \multicolumn{1}{c|}{18}          & \multicolumn{1}{c|}{69.87}          & \multicolumn{1}{c|}{46.14}          & \multicolumn{1}{c|}{56.40}          & 23.73          & \multicolumn{1}{c|}{18}          & \multicolumn{1}{c|}{70.14}          & \multicolumn{1}{c|}{46.61}          & \multicolumn{1}{c|}{55.31}       & 23.53                      & \multicolumn{1}{c|}{18}          & \multicolumn{1}{c|}{\underline{81.07}}    & \multicolumn{1}{c|}{59.13}          & \multicolumn{1}{c|}{69.26}          & 21.94          & \begin{tabular}[c]{@{}p{4.5cm}@{}}  Proposed a semantics-driven generative replay framework\end{tabular}                                                                                                                                           \\ \cline{2-20} 
                                     & Data Replay                                                                              & ERDFR~\cite{liu2022few}                             & ECCV 22                           & \multicolumn{1}{c|}{18}          & \multicolumn{1}{c|}{71.84}          & \multicolumn{1}{c|}{48.21}          & \multicolumn{1}{c|}{58.02}          & 23.63          & \multicolumn{1}{c|}{20}          & \multicolumn{1}{c|}{74.40}          & \multicolumn{1}{c|}{50.14}          & \multicolumn{1}{c|}{60.78}       & 24.26                      & \multicolumn{1}{c|}{18}          & \multicolumn{1}{c|}{75.90}          & \multicolumn{1}{c|}{52.39}          & \multicolumn{1}{c|}{61.52}          & 23.51          & \begin{tabular}[c]{@{}p{4.5cm}@{}}  Proved the effectiveness of DR in FSCIL and proposed a data-free replay method\end{tabular}                                                                                                           \\ \cline{2-20} 
                                     & \begin{tabular}[c]{@{}l@{}} Data Replay\\ Knowledge Distillation\end{tabular}             & FDD~\cite{zhu2022feature}                               & PRAI 22                            & \multicolumn{1}{c|}{18}          & \multicolumn{1}{c|}{64.14}          & \multicolumn{1}{c|}{42.01}          & \multicolumn{1}{c|}{52.44}          & 22.13          & \multicolumn{1}{c|}{18}          & \multicolumn{1}{c|}{64.94}          & \multicolumn{1}{c|}{41.73}          & \multicolumn{1}{c|}{52.52}       & 23.21                      & \multicolumn{1}{c|}{18}          & \multicolumn{1}{c|}{-}              & \multicolumn{1}{c|}{-}              & \multicolumn{1}{c|}{-}              & -              & \begin{tabular}[c]{@{}p{4.5cm}@{}} Fixed the shallow layers and fine-tune the deep layers with KD loss and CE loss\end{tabular}                                                                                                          \\ \cline{2-20} 
                                     & Pseudo Scenarios                                                                         & SPPR~\cite{zhu2021self}                              & CVPR 21                            & \multicolumn{1}{c|}{18}          & \multicolumn{1}{c|}{61.45}          & \multicolumn{1}{c|}{41.92}          & \multicolumn{1}{c|}{52.76}          & 19.53          & \multicolumn{1}{c|}{18}          & \multicolumn{1}{c|}{63.97}          & \multicolumn{1}{c|}{43.32}          & \multicolumn{1}{c|}{54.45}       & 20.65                      & \multicolumn{1}{c|}{18}          & \multicolumn{1}{c|}{68.68}          & \multicolumn{1}{c|}{37.33}          & \multicolumn{1}{c|}{49.32}          & 31.35          & \begin{tabular}[c]{@{}p{4.5cm}@{}} Built a randomly episodic training based on a novel self-promoted prototype refinement mechanism\end{tabular}                                                                                                  \\ \cline{2-20} 
                                     & Pseudo Scenarios                                                                         & FACT~\cite{zhou2022forward}                              & CVPR 22                            & \multicolumn{1}{c|}{18}          & \multicolumn{1}{c|}{72.56}          & \multicolumn{1}{c|}{50.49}          & \multicolumn{1}{c|}{60.70}          & 22.07          & \multicolumn{1}{c|}{20}          & \multicolumn{1}{c|}{74.60}          & \multicolumn{1}{c|}{52.10}          & \multicolumn{1}{c|}{62.24}       & 22.50                      & \multicolumn{1}{c|}{18}          & \multicolumn{1}{c|}{75.90}          & \multicolumn{1}{c|}{56.94}          & \multicolumn{1}{c|}{64.42}          & 18.96          & \begin{tabular}[c]{@{}p{4.5cm}@{}} Proposed a forward compatible training strategy, which reserves embedding space for new classes during base learning\end{tabular}                                                                              \\ \cline{2-20} 
                                     & \begin{tabular}[c]{@{}l@{}} Pseudo Scenarios\\  Representation Learning\end{tabular}       & ALICE~\cite{peng2022few}                             & ECCV 22                            & \multicolumn{1}{c|}{18}          & \multicolumn{1}{c|}{\underline{80.60}}    & \multicolumn{1}{c|}{\underline{55.70}}    & \multicolumn{1}{c|}{\underline{ 63.99}}    & 24.90          & \multicolumn{1}{c|}{18}          & \multicolumn{1}{c|}{79.00}          & \multicolumn{1}{c|}{54.10}          & \multicolumn{1}{c|}{63.21}       & 24.90                      & \multicolumn{1}{c|}{18}          & \multicolumn{1}{c|}{77.40}          & \multicolumn{1}{c|}{60.10}          & \multicolumn{1}{c|}{65.75}          & \underline{\textbf{17.30}}    & \begin{tabular}[c]{@{}p{4.5cm}@{}} ALICE used angular penalty loss for discriminated and generalized feature learning\end{tabular}                                                                                                         \\ \cline{2-20} 
                                     & \begin{tabular}[c]{@{}l@{}} Pseudo Scenarios\\ Representation Learning\end{tabular}       & SAVC~\cite{song2023learning}                              & CVPR 23                            & \multicolumn{1}{c|}{18}          & \multicolumn{1}{c|}{ \underline{\textbf{81.12}}} & \multicolumn{1}{c|}{\underline{\textbf{57.11}}} & \multicolumn{1}{c|}{\underline{\textbf{67.05}}} & 24.01          & \multicolumn{1}{c|}{20}          & \multicolumn{1}{c|}{78.77}          & \multicolumn{1}{c|}{53.12}          & \multicolumn{1}{c|}{63.63}       & 25.65                      & \multicolumn{1}{c|}{18}          & \multicolumn{1}{c|}{\underline{\textbf{81.85}}} & \multicolumn{1}{c|}{\underline{62.50}}           & \multicolumn{1}{c|}{69.35}          & 19.35          & \begin{tabular}[c]{@{}p{4.5cm}@{}} Used supervised contrast learning and virtual classes to initialize backbone so that it can have good generalization performance\end{tabular}                                                                                  \\ \hline
\multirow{3}{*}{\rotatebox{90}{Structure-based}}     & Dynamic Structure                                                                        & TOPIC~\cite{tao2020few}                             & CVPR 20                            & \multicolumn{1}{c|}{18}          & \multicolumn{1}{c|}{61.31}          & \multicolumn{1}{c|}{24.42}          & \multicolumn{1}{c|}{39.64}          & 36.89          & \multicolumn{1}{c|}{18}          & \multicolumn{1}{c|}{64.10}          & \multicolumn{1}{c|}{29.37}          & \multicolumn{1}{c|}{42.62}       & 34.73                      & \multicolumn{1}{c|}{18}          & \multicolumn{1}{c|}{68.68}          & \multicolumn{1}{c|}{26.28}          & \multicolumn{1}{c|}{43.92}          & 42.40          & \begin{tabular}[c]{@{}p{4.5cm}@{}} Introduced FSCIL setting for the first time, and proposed TOPIC framework based on neural gas network\end{tabular}                                                                                             \\ \cline{2-20} 
                                     & \begin{tabular}[c]{@{}l@{}}  Dynamic Structure\\  Pseudo Scenarios\\ Attention\end{tabular} & CEC~\cite{zhang2021few}                              & CVPR 21                            & \multicolumn{1}{c|}{18}          & \multicolumn{1}{c|}{72.00}          & \multicolumn{1}{c|}{47.63}          & \multicolumn{1}{c|}{57.75}          & 24.37          & \multicolumn{1}{c|}{20}          & \multicolumn{1}{c|}{73.07}          & \multicolumn{1}{c|}{49.14}          & \multicolumn{1}{c|}{59.53}       & 23.93                      & \multicolumn{1}{c|}{18}          & \multicolumn{1}{c|}{75.85}          & \multicolumn{1}{c|}{52.28}          & \multicolumn{1}{c|}{61.33}          & 23.57          & \begin{tabular}[c]{@{}p{4.5cm}@{}} Proposed CEC framework based on GAT to propagate context information, and it was trained pseudo-incremental sessions\end{tabular}                                                                               \\ \cline{2-20} 
                                     & Dynamic Structure                                                                        & DSN~\cite{yang2022dynamic}                              & TPAMI 22                           & \multicolumn{1}{c|}{18}          & \multicolumn{1}{c|}{66.95}          & \multicolumn{1}{c|}{45.89}          & \multicolumn{1}{c|}{54.39}          & 21.06          & \multicolumn{1}{c|}{18}          & \multicolumn{1}{c|}{73.00}          & \multicolumn{1}{c|}{50.00}          & \multicolumn{1}{c|}{60.14}       & 23.00                      & \multicolumn{1}{c|}{18}          & \multicolumn{1}{c|}{80.86}          & \multicolumn{1}{c|}{\underline{\textbf{63.21}}}    & \multicolumn{1}{c|}{\underline{\textbf{71.02}}} & \underline{17.65}          & \begin{tabular}[c]{@{}p{4.5cm}@{}} Proposed DSN, an adaptively updating network with compressive node expansion to “support” the feature space\end{tabular}                                             \\ \hline
\multirow{16}{*}{\rotatebox{90}{Optimization-based}} & Representation Learning                                                                  & SFbFSCIL~\cite{cheraghian2021synthesized}                         & ICCV 21                            & \multicolumn{1}{c|}{18}          & \multicolumn{1}{c|}{61.37}          & \multicolumn{1}{c|}{42.23}          & \multicolumn{1}{c|}{50.73}          & \underline{19.14}    & \multicolumn{1}{c|}{18}          & \multicolumn{1}{c|}{62.00}          & \multicolumn{1}{c|}{41.69}          & \multicolumn{1}{c|}{50.86}       & \underline{20.31}                & \multicolumn{1}{c|}{18}          & \multicolumn{1}{c|}{68.78}          & \multicolumn{1}{c|}{43.23}          & \multicolumn{1}{c|}{51.84}          & 25.55          & \begin{tabular}[c]{@{}p{4.5cm}@{}} Proposed the use of a mixture of subspaces and synthesized features by VAE to reduce the forgetting and overfitting problem\end{tabular}                                                                       \\ \cline{2-20} 
                                     & Representation Learning            & FSLL+SS~\cite{mazumder2021few}                           & AAAI 21                            & \multicolumn{1}{c|}{18}          & \multicolumn{1}{c|}{68.85}          & \multicolumn{1}{c|}{43.92}          & \multicolumn{1}{c|}{53.92}          & 24.93          & \multicolumn{1}{c|}{18}          & \multicolumn{1}{c|}{66.76}          & \multicolumn{1}{c|}{39.57}          & \multicolumn{1}{c|}{48.71}       & 27.19                      & \multicolumn{1}{c|}{18}          & \multicolumn{1}{c|}{75.63}          & \multicolumn{1}{c|}{55.82}          & \multicolumn{1}{c|}{62.62}          & 19.81          & \begin{tabular}[c]{@{}p{4.5cm}@{}} FSLL trained only a few selected parameters to limit overfitting, and leverages self-supervision\end{tabular}                                                                                                   \\ \cline{2-20} 
                                     & Representation Learning                                                                  & MgSvF~\cite{zhao2021mgsvf}                             & TPAMI 21                           & \multicolumn{1}{c|}{18}          & \multicolumn{1}{c|}{63.09}          & \multicolumn{1}{c|}{45.76}          & \multicolumn{1}{c|}{52.87}          & \underline{\textbf{17.33}} & \multicolumn{1}{c|}{18}          & \multicolumn{1}{c|}{74.24}          & \multicolumn{1}{c|}{51.40}          & \multicolumn{1}{c|}{61.67}       & 22.84                      & \multicolumn{1}{c|}{18}          & \multicolumn{1}{c|}{72.29}          & \multicolumn{1}{c|}{54.33}          & \multicolumn{1}{c|}{62.37}          & 17.96          & \begin{tabular}[c]{@{}p{4.5cm}@{}} MgSvF used frequency-aware regularization and feature space composition to balance old and new knowledge\end{tabular}                                                                                      \\ \cline{2-20} 
                                     & Representation Learning                                                                  & CLOM~\cite{zou2022margin}                              & NeurIPS 22                            & \multicolumn{1}{c|}{18}          & \multicolumn{1}{c|}{73.08}          & \multicolumn{1}{c|}{48.00}          & \multicolumn{1}{c|}{58.48}          & 25.08          & \multicolumn{1}{c|}{20}          & \multicolumn{1}{c|}{74.20}          & \multicolumn{1}{c|}{50.25}          & \multicolumn{1}{c|}{60.57}       & 23.95                      & \multicolumn{1}{c|}{18}          & \multicolumn{1}{c|}{79.57}          & \multicolumn{1}{c|}{59.58}          & \multicolumn{1}{c|}{67.17}          & 19.99          & \begin{tabular}[c]{@{}p{4.5cm}@{}} Interpreted the dilemma of the margin-based classification as a class-level overfitting problem and proposed CLOM to mitigate it\end{tabular}                                                                  \\ \cline{2-20} 
                                     & Representation Learning                                                                  & RE~\cite{yao2022few}                                & JEI 22                             & \multicolumn{1}{c|}{18}          & \multicolumn{1}{c|}{70.74}          & \multicolumn{1}{c|}{45.48}          & \multicolumn{1}{c|}{56.30}          & 25.26          & \multicolumn{1}{c|}{18}          & \multicolumn{1}{c|}{70.72}          & \multicolumn{1}{c|}{47.52}          & \multicolumn{1}{c|}{57.77}       & 23.20                      & \multicolumn{1}{c|}{18}          & \multicolumn{1}{c|}{-}              & \multicolumn{1}{c|}{-}              & \multicolumn{1}{c|}{-}              & -              & \begin{tabular}[c]{@{}p{4.5cm}@{}} Proposed representation enhancement method by exploring correlations with previously learned classes\end{tabular}                                                                                              \\ \cline{2-20} 
                                     & Representation Learning                                                                  & WaRP~\cite{kim2023warping}                              & ICLR 23                            & \multicolumn{1}{c|}{18}          & \multicolumn{1}{c|}{72.99}          & \multicolumn{1}{c|}{50.65}          & \multicolumn{1}{c|}{59.69}          & 22.34          & \multicolumn{1}{c|}{20}          & \multicolumn{1}{c|}{\underline{\textbf{80.31}}} & \multicolumn{1}{c|}{\underline{54.74}}          & \multicolumn{1}{c|}{\underline{65.82}}       & 25.57                      & \multicolumn{1}{c|}{18}          & \multicolumn{1}{c|}{77.74}          & \multicolumn{1}{c|}{57.01}          & \multicolumn{1}{c|}{64.66}          & 20.73          & \begin{tabular}[c]{@{}p{4.5cm}@{}} WaRP, a weight space rotation process that compressed old knowledge into key parameters, allowing fine-tuning without forgetting\end{tabular}                                                                  \\ \cline{2-20} 
                                     & Representation Learning                     & TEEN~\cite{wang2023few2}                              & NeurIPS 23                            & \multicolumn{1}{c|}{18}          & \multicolumn{1}{c|}{73.53}          & \multicolumn{1}{c|}{52.08}          & \multicolumn{1}{c|}{61.45} & 21.45          & \multicolumn{1}{c|}{20}          & \multicolumn{1}{c|}{74.92}          & \multicolumn{1}{c|}{52.64}          & \multicolumn{1}{c|}{63.10}        & 22.28                      & \multicolumn{1}{c|}{18}          & \multicolumn{1}{c|}{77.26} & \multicolumn{1}{c|}{59.31}          & \multicolumn{1}{c|}{66.63}          & 18.13          & \begin{tabular}[c]{@{}p{4.5cm}@{}} TEEN, a training-free calibration strategy, enhanced new class discriminability by fusing new and weighted base class prototypes.\end{tabular}       \\ \cline{2-20}

                                     & \begin{tabular}[c]{@{}l@{}} Knowledge Distillation\\ Attention\end{tabular}               & SaKD~\cite{cheraghian2021semantic}                              & CVPR 21                            & \multicolumn{1}{c|}{18}          & \multicolumn{1}{c|}{61.33}          & \multicolumn{1}{c|}{38.73}          & \multicolumn{1}{c|}{48.20}          & 22.60          & \multicolumn{1}{c|}{18}          & \multicolumn{1}{c|}{64.03}          & \multicolumn{1}{c|}{34.94}          & \multicolumn{1}{c|}{45.53}       & 29.09                      & \multicolumn{1}{c|}{18}          & \multicolumn{1}{c|}{68.23}          & \multicolumn{1}{c|}{32.96}          & \multicolumn{1}{c|}{46.13}          & 35.27          & \begin{tabular}[c]{@{}p{4.5cm}@{}} Proposed a semantic-aware distillation method and an attention driven alignment strategy to mitigate catastrophic forgetting\end{tabular}                                                                       \\ \cline{2-20} 
                                     & Knowledge Distillation                                                                   & ERL++~\cite{dong2021few}                             & AAAI 21                            & \multicolumn{1}{c|}{18}          & \multicolumn{1}{c|}{61.71}          & \multicolumn{1}{c|}{40.77}          & \multicolumn{1}{c|}{49.84}          & 20.94          & \multicolumn{1}{c|}{18}          & \multicolumn{1}{c|}{73.70}          & \multicolumn{1}{c|}{48.25}          & \multicolumn{1}{c|}{59.31}       & 25.45                      & \multicolumn{1}{c|}{18}          & \multicolumn{1}{c|}{73.52}          & \multicolumn{1}{c|}{52.28}          & \multicolumn{1}{c|}{61.18}          & 21.24          & \begin{tabular}[c]{@{}p{4.5cm}@{}} Proposed the exemplar relation distillation and degree-based graph construction method to model the exemplar relationship\end{tabular}                                                                         \\ \cline{2-20} 
                                     & \begin{tabular}[c]{@{}l@{}} Knowledge Distillation\\ Attention\end{tabular}               & BiDistFSCIL~\cite{zhao2023few}                       & CVPR 23                            & \multicolumn{1}{c|}{18}          & \multicolumn{1}{c|}{74.65}          & \multicolumn{1}{c|}{52.22}          & \multicolumn{1}{c|}{61.42}          & 22.43          & \multicolumn{1}{c|}{18}          & \multicolumn{1}{c|}{\underline{79.45}}    & \multicolumn{1}{c|}{\underline{\textbf{55.88}}}    & \multicolumn{1}{c|}{\underline{\textbf{66.14}}} & 23.57                      & \multicolumn{1}{c|}{18}          & \multicolumn{1}{c|}{79.12}          & \multicolumn{1}{c|}{60.93}          & \multicolumn{1}{c|}{67.34}          & 18.19          & \begin{tabular}[c]{@{}p{4.5cm}@{}} It proposed a KD strategy with two teacher models, designed a two-branch network, and used an attention mechanism to aggregate the predictions from both branches.\end{tabular} \\ 
                                     \cline{2-20} 
                                     & \begin{tabular}[c]{@{}l@{}} Meta Learning\\  Attention\end{tabular}                        & MetaFSCIL~\cite{chi2022metafscil}                         & CVPR 22                            & \multicolumn{1}{c|}{18}          & \multicolumn{1}{c|}{72.04}          & \multicolumn{1}{c|}{49.19}          & \multicolumn{1}{c|}{58.85}          & 22.85          & \multicolumn{1}{c|}{20}          & \multicolumn{1}{c|}{74.50}          & \multicolumn{1}{c|}{49.97}          & \multicolumn{1}{c|}{60.79}       & 24.53                      & \multicolumn{1}{c|}{18}          & \multicolumn{1}{c|}{75.90}          & \multicolumn{1}{c|}{52.64}          & \multicolumn{1}{c|}{61.93}          & 23.26          & \begin{tabular}[c]{@{}p{4.5cm}@{}} Adopted a bi-level meta-learning optimization and bi-directional guided modulation approach\end{tabular}                                                                                                       \\ \cline{2-20} 
                                     &  Meta Learning                                                                            & CSR~\cite{zheng2021few}                               & ICDMW 21                           & \multicolumn{1}{c|}{18}          & \multicolumn{1}{c|}{67.67}          & \multicolumn{1}{c|}{44.52}          & \multicolumn{1}{c|}{54.11}          & 23.15          & \multicolumn{1}{c|}{18}          & \multicolumn{1}{c|}{72.02}          & \multicolumn{1}{c|}{49.00}          & \multicolumn{1}{c|}{59.07}       & 23.02                      & \multicolumn{1}{c|}{18}          & \multicolumn{1}{c|}{74.69}          & \multicolumn{1}{c|}{55.09}          & \multicolumn{1}{c|}{62.32}          & 19.60          & \begin{tabular}[c]{@{}p{4.5cm}@{}} Introduced class structures and adopted an alignment kernel, employing meta-learning process\end{tabular}                                                                                                      \\ \cline{2-20} 
                                     & \begin{tabular}[c]{@{}l@{}}\tiny Meta Learning\\ \tiny Attention\end{tabular}                        & LIMIT~\cite{zhou2022few}                             & TPAMI 22                           & \multicolumn{1}{c|}{18}          & \multicolumn{1}{c|}{72.32}          & \multicolumn{1}{c|}{49.19}          & \multicolumn{1}{c|}{59.06}          & 23.13          & \multicolumn{1}{c|}{20}          & \multicolumn{1}{c|}{73.81}          & \multicolumn{1}{c|}{51.23}          & \multicolumn{1}{c|}{61.84}       & 22.58                      & \multicolumn{1}{c|}{18}          & \multicolumn{1}{c|}{75.89}          & \multicolumn{1}{c|}{57.41}          & \multicolumn{1}{c|}{65.48}          & 18.48          & \begin{tabular}[c]{@{}p{4.5cm}@{}} LIMIT, a meta-learning based paradigm, which synthesized fake tasks to build a generalizable feature space for unseen tasks\end{tabular}                                                                       \\ \cline{2-20} 
                                     & Others                                                                                   & F2M~\cite{shi2021overcoming}                               & NeurIPS 21                            & \multicolumn{1}{c|}{18}          & \multicolumn{1}{c|}{67.28}          & \multicolumn{1}{c|}{44.65}          & \multicolumn{1}{c|}{54.89}          & 22.63          & \multicolumn{1}{c|}{18}          & \multicolumn{1}{c|}{64.71}          & \multicolumn{1}{c|}{44.67}          & \multicolumn{1}{c|}{53.65}       & \underline{\textbf{20.04}}             & \multicolumn{1}{c|}{18}          & \multicolumn{1}{c|}{\underline{81.07}}    & \multicolumn{1}{c|}{60.26}          & \multicolumn{1}{c|}{\underline{69.49}}    & 20.81          & \begin{tabular}[c]{@{}p{4.5cm}@{}} Proposed a method by finding flat local minima during base training and fine-tuning within this region when learning new classes\end{tabular}                                                                  \\ 
                                     \hline
\end{tabular}
\label{tab3}
\end{table*}

For FSCIC methods, the performance of the backbone achieved on the base session is crucial for subsequent IL. From the analysis of SA across the three datasets in Tab.~\ref{tab3}, it can be seen that the top five methods on \emph{mini}ImageNet are: SAVC~\cite{song2023learning} (81.12\%), ALICE~\cite{peng2022few} (80.60\%), BiDistFSCIL~\cite{zhao2023few} (74.65\%), TEEN~\cite{wang2023few2} (73.53\%), and CLOM~\cite{zou2022margin} (73.08\%). On CIFAR-100, the top five methods are: WaRP~\cite{kim2023warping} (80.31\%), BiDistFSCIL (79.45\%), ALICE (79.00\%), SAVC (78.77\%), and TEEN (74.92\%). On the CUB-200 dataset, the top five methods are: SAVC (81.85\%), F2M~\cite{shi2021overcoming} and FSIL-GAN~\cite{agarwal2022semantics} (81.07\%), DSN~\cite{yang2022dynamic} (80.86\%), and BiDistFSCIL (79.12\%). It is noteworthy that the SAVC, based on virtual class synthesis, achieved the best initial performance on \emph{mini}ImageNet and CUB-200; the ALICE framework, which leverages metric learning and pseudo-data synthesis, also performed exceptionally well on these datasets. This indicates that the virtual class strategy is highly effective in improving performance in the initial session. Apart from that, it has been found that almost all methods that achieved top five performance on the base session also achieved top five performance in terms of the AA index. It reflects the influence of the performance achieved on the base session.

The performance obtained in the final session reflects the learning capability of the FSCIC model for incremental classes and the stability of keeping old knowledge. However, as the model learned in each incremental session will be tested on all seen classes, and the number of classes involved in the base session is large, the performance obtained on each session cannot fully represent the model's IL ability. In contrast, the PD value can better reflect the model's ability to resist forgetting. In Tab.~\ref{tab3}, the top five methods with the lowest PD values on \emph{mini}ImageNet are: MgSvF~\cite{zhao2021mgsvf} (17.33\%), SFbFSCIL~\cite{cheraghian2021synthesized} (19.14\%), SPPR~\cite{zhu2021self} (19.53\%), ERL++~\cite{dong2021few} (20.94\%), and DSN (21.06\%). On CIFAR-100, the top five methods are: F2M~\cite{shi2021overcoming} (20.04\%), SFbFSCIL (20.31\%), SPPR (20.65\%), TEEN (22.28\%), and FACT~\cite{zhou2022forward} (22.50\%). On CUB-200, the top five methods are: ALICE (17.30\%), DSN (17.65\%), MgSvF (17.96\%), TEEN (18.13\%), and BiDistFSCIL(18.19\%). It can be seen that SPPR, which constructs pseudo incremental sessions, and SFbFSCIL, which is based on feature space fusion and VAE feature synthesis, both have achieved good PD values on the first two datasets. DSN, based on dynamic network structure, and MgSvF, based on frequency domain analysis, performed well on the first and last datasets. Furthermore, combining the performance of other methods on the three datasets, it can be found that techniques such as KD, pseudo-incremental scenario construction, dynamic structures, and feature optimization can effectively alleviate the catastrophic forgetting problem.

\subsubsection{Main Issues and Facts}
In FSCIC, the current issues primarily encompass a lack of comprehensive evaluation metrics, unfairness in experimental conditions, and inconsistencies with real-world scenarios. Most studies use AA or PD values to measure model performance, but they can not reflect the performance details during the continuous learning process~\cite{peng2022few}. Furthermore, the variability in choosing backbone networks and the introduction of additional data introduce inherent biases when comparing different methodologies. Most importantly, the current setting of FSCIC faces challenges in real-world implementation.

\section{Few-shot Class-incremental Object Detection}
\label{sec:FSCIOD}
Since the instance segmentation framework in FSCIL generally has object detection capabilities, this section discusses them together. Firstly, the difference with FSCIC is presented. Then, existing methods are systematically summarized from the perspectives of anchor-free and anchor-based frameworks. Finally, the paper summarizes the entire work, including performance comparisons and discussions of key issues.

\subsection{Difference with Classification}
In contrast to the classification task in FSCIL, FSCIOD aims to enable the model to continuously learn new classes from limited samples while achieving accurate localization (using bounding box regression or segmentation) and classification of each corresponding individual object in an image~\cite{ganea2021incremental,yin2022sylph,dong2023incremental}. The model is also required to retain the capability of object localization and classification for the old classes.

Similar to the classification setting in FSCIL provided in Sec.~\ref{sec:over:pd}, the training data for FSCIOD can be divided into the base and new training sets. However, there is a difference. In the classification task, the new classes are typically further divided into multiple incremental sessions in the form of $N-$way $K-$shot, while in the current object detection setting, the new classes usually are formed as one incremental session. Specifically, the training sets for FSCIOD can be denoted as $\{D^b_{train}, D^n_{train}\}$, where the base training set $D^b_{train}$ contains a large number of labeled training samples and can be represented as $D^b_{train}={(x_i,y_i)}^{n_0}{i=1}$, where $x_i$, $y_i$, and $n_0$ represent the training sample, its corresponding ground truth set, and the number of base samples, respectively. Similar to the classification setting, the new training set $D^n_{train}=\{(x_i,y_i)\}^{N\times K}_{i=1}$ is in the form of $N-$way $K-$shot. Note that the classes in the base and new training sets do not intersect. The evaluation process for the object detection task in FSCIL is similar to the classification task. After learning the new training set, the model is evaluated on the performance of all seen classes, \emph{i.e.}, the union of testing data from all seen classes.

It is important to note that in incremental images, even if a single image contains multiple objects of different classes, only the ground truth set for the current class is provided to align with the few-shot class-incremental setup.

\subsection{Methods}
FSCIOD requires simultaneously localizing and classifying new class objects during IL while not forgetting the old knowledge. This poses a greater challenge compared to classification in FSCIL. Current methods include both anchor-based and anchor-free frameworks. Generally, anchor-based detectors have superior detection performance, but they suffer from lower efficiency and flexibility due to the design of anchors. On the other hand, anchor-free detectors are more efficient and flexible.

\subsubsection{Anchor-free Frameworks}
\label{sec:FSCIOD:aff}
Recently, some studies~\cite{perez2020incremental,cheng2021meta,yin2022sylph,feng2022incremental,dong2023incremental} have adopted anchor-free frameworks to perform this task. The reason is that anchor-free frameworks can effectively handle incremental classes without defining anchor boxes. According to their detection framework, these studies can be classified into three categories: CentreNet-based, FCOS-based, and DETR-based methods.

\textbf{CentreNet-based Methods:}
CentreNet~\cite{zhou2019objects} redefined object detection as a \emph{point$+$attribute} regression problem. During detection, it divided the input image into different regions, each with a centre point. CentreNet made predictions to determine whether the centre point corresponds to an object. Then, it predicted the class and confidence for this object. CentreNet also adjusted the centre point to obtain the accurate location and regressed the object's width and height. By maintaining independent prediction heatmaps for each class and using activation thresholding for independent object detection, CentreNet supported incremental registration of new classes. Based on CentreNet, \emph{Perez-Rua et al.}~\cite{perez2020incremental} proposed the ONCE framework, which incorporated meta-learning for object detection in FSCIL. It decomposed CentreNet into a fixed universal feature extractor trained on base classes and a meta-learned object localizer with class-specific parameters. In the few-shot incremental detection scenario, the model only required forward propagation for registration without model updating or accessing base data. Additionally, \emph{Cheng et al.}~\cite{cheng2021meta} also utilized CentreNet as the backbone and introduced meta-learning based on MAML~\cite{finn2017model}. First, meta-learning provided good initialization for the object localizer based on base data, enabling easy fine-tuning with few-shot samples from new classes. Furthermore, the filter parameters of base classes were retained. The meta-learner determined the remaining parameters of the object localizer. The study also concluded that the main factor limiting the performance of new classes is the overfitting of the feature extractor to base classes, resulting in insufficient generalization.

\textbf{FCOS-based Methods:}
Similarly, recent works have adopted it as a backbone due to the strong performance and class-agnostic localization capability of FCOS~\cite{tian2019fcos}. For instance, Sylph proposed by \emph{Yin et al.}~\cite{yin2022sylph} decomposed the detection framework into a class-agnostic detector and a novel classifier to enable continual learning of new classes. Specifically, FCOS was employed in Sylph for class-agnostic object localization. Since optimizing softmax can lead to catastrophic forgetting~\cite{yin2022sylph,farquhar2018towards}, Sylph replaced it with multiple binary sigmoid-based classifiers, each independently handling its own set of parameters. When adding new classes, a new set of classifier parameters can be generated with zero interference between predictions of different classes. In addition, \emph{Feng et al.}~\cite{feng2022incremental} proposed two modules inspired by the phenomenon of establishing new connections between memory cells in the brain when new memories appear. The first was called the MCH module, which added a classification branch to predict new classes each time they appeared. The second was called the BPMCH module, which added a new backbone that was initialized with the weights of the base class backbone to transfer more knowledge from the base classes to the new classes. In this work, FCOS and ATSS~\cite{zhang2020bridging} were employed as the baseline detectors. Training started on the base classes and was then fine-tuned on the new classes, ensuring the retention of knowledge learned from the base classes and transferring that knowledge to the new classes.

\textbf{DETR-based Method:} In anchor-free frameworks, in addition to the methods based on CentreNet and FCOS, another work adopts the DETR framework~\cite{carion2020end} as the backbone. Specifically, \emph{Dong et al.}~\cite{dong2023incremental} proposed the incremental-DETR, which firstly introduced DETR to FSCIOD. This method consisted of two stages: First, the entire network was pre-trained using a large amount of data from the base classes, and the class-specific components of DETR (including the projection layer and classification head for specific classes) were fine-tuned using self-supervision from additional object proposals generated by selective search algorithm~\cite{uijlings2013selective} as pseudo labels. Then, the CNN backbone, transformer, and regression head were fixed, and an incremental few-shot fine-tuning strategy was introduced to fine-tune and distill knowledge from the class-specific components of DETR. This strategy encouraged the framework to detect new classes without catastrophic forgetting.

\subsubsection{Anchor-based Frameworks}
\label{sec:FSCIOD:abf}
In addition to anchor-free frameworks, there have been some studies~\cite{ganea2021incremental,nguyen2022ifs} that adopt the anchor-based framework, Mask R-CNN~\cite{he2017mask}, to address object detection and instance segmentation in FSCIL. Mask R-CNN is a popular framework for the instance segmentation, which extended the Faster R-CNN~\cite{ren2015faster} architecture by incorporating a mask prediction branch. It is a two-stage approach that combines object detection and pixel-level segmentation into one framework. Currently, there is limited research on instance segmentation in FSCIL, and all utilize Mask R-CNN as the backbone. For example, \emph{Ganea et al.}~\cite{ganea2021incremental} proposed the iMTFA framework while initially introducing the setting of few-shot incremental instance segmentation. Specifically, they added an instance segmentation branch (similar to Mask R-CNN to Faster R-CNN) to the few-shot object detection framework TFA~\cite{wang2020frustratingly}, resulting in MTAF. One drawback of MTAF was that it required continual fine-tuning when adding new classes. Thus, they extended MTFA to an incremental method called iMTFA. In this framework, the regression and mask prediction heads were class-agnostic. Additionally, the framework learned a feature extractor that generates discriminative features. The feature extractor was used for new classes to compute the averaged prototype vectors for each class, which were then concatenated with the existing classifier. This enabled few-shot incremental instance segmentation without the need for further training. Furthermore, \emph{Nguyen and Todorovic}~\cite{nguyen2022ifs} extended the Mask R-CNN framework in the second stage: a new object class classifier based on the probit function~\cite{spiegelhalter1990sequential} and a new uncertainty-guided bounding box predictor. The former utilized Bayesian learning to address the scarcity of training examples for new classes. The latter not only predicted object bounding boxes but also estimated the uncertainty of the predictions, which guided the refinement of bounding boxes. Two new loss functions were also specified based on the estimated object-class distribution and bounding-box uncertainty.

\subsection{Summary}
\subsubsection{Performance Comparison}
In this section, we summarize the performance of FSCIOD methods. We summarize the performance of relevant methods on COCO and VOC in Tab.~\ref{tab4}. To fully elucidate the attributes of each method, Tab.~\ref{tab4} includes their types and specific taxonomy categories. In addition, the backbone employed by each method is furnished in this table. Because some methods can achieve object detection and instance segmentation simultaneously, we have added a "task" column in Tab.~\ref{tab4} to denote performance on related tasks. FSCIOD methods are evaluated in two ways: standard evaluation on COCO and cross-dataset evaluation on COCO and VOC. Tab.~\ref{tab4} presents mAP, mAP50, and mAR values, along with method highlights.

\begin{table*}[htp]
\scriptsize
\centering
\renewcommand{\arraystretch}{1}
\setlength\tabcolsep{1pt}
\caption{The performance of FSCIOD methods. The data are extracted from the original papers. The taxonomy is abbreviated to taxo. We use “-” to mark the dataset without reporting in the original papers. (In \%)} 
\vspace{-10pt}
\begin{tabular}{|c|c|c|c|c|c|c|ccccccccc|ccc|c|}
\hline
\multirow{3}{*}{\textbf{Type}} & \multirow{3}{*}{\textbf{Taxo.}}                                           & \multirow{3}{*}{\textbf{Method}}                                             & \multirow{3}{*}{\textbf{Venue}} & \multirow{3}{*}{\textbf{\begin{tabular}[c]{@{}c@{}}Backbone\\(ResNet)\end{tabular}}} & \multirow{3}{*}{\textbf{Task}} & \multirow{3}{*}{\textbf{Shot}} & \multicolumn{9}{c|}{\textbf{COCO}}                                                                                                                                                                                                                                                                        & \multicolumn{3}{c|}{\textbf{VOC}}                                                   & \multirow{3}{*}{\textbf{Highlights}}                                                                                   \\ \cline{8-19}
                               &                                                                              &                                                                              &                        &                                                                              &                       &                       & \multicolumn{3}{c|}{\textbf{Base}}                                                                       & \multicolumn{3}{c|}{\textbf{Novel}}                                                                      & \multicolumn{3}{c|}{\textbf{Overall}}                                               & \multicolumn{3}{c|}{\textbf{Novel}}                                                 &                                                                                                                                                                                                    \\ \cline{8-19}
                               &                                                                              &                                                                              &                        &                                                                              &                       &                       & \multicolumn{1}{c|}{\textbf{mAP}} & \multicolumn{1}{c|}{\textbf{mAP50}} & \multicolumn{1}{c|}{\textbf{mAR}} & \multicolumn{1}{c|}{\textbf{mAP}} & \multicolumn{1}{c|}{\textbf{mAP50}} & \multicolumn{1}{c|}{\textbf{mAR}} & \multicolumn{1}{c|}{\textbf{mAP}} & \multicolumn{1}{c|}{\textbf{mAP50}} & \textbf{mAR} & \multicolumn{1}{c|}{\textbf{mAP}} & \multicolumn{1}{c|}{\textbf{mAP50}} & \textbf{mAR} &                                                                                                                                                                                                    \\ \hline
\multirow{21}{*}{\rotatebox{90}{Anchor-free}}  & \multirow{9}{*}{\rotatebox{90}{CentreNet-based}}                                             & \multirow{3}{*}{ONCE~\cite{perez2020incremental}}                                                        & \multirow{3}{*}{CVPR 20}  & \multirow{3}{*}{\textbf{50}}                                                 & \multirow{3}{*}{D}    & \textbf{1}            & \multicolumn{1}{c|}{17.90}       & \multicolumn{1}{c|}{-}             & \multicolumn{1}{c|}{19.50}       & \multicolumn{1}{c|}{0.70}        & \multicolumn{1}{c|}{-}             & \multicolumn{1}{c|}{6.30}        & \multicolumn{1}{c|}{13.60}       & \multicolumn{1}{c|}{-}             & 16.20       & \multicolumn{1}{c|}{-}           & \multicolumn{1}{c|}{-}             & -           & \multirow{3}{*}{\begin{tabular}[c]{@{}p{4cm}@{}}Proposed FSCIOD setting and introduced the first work, ONCE\end{tabular}}                                                                                                                       \\ \cline{7-19}
                               &                                                                              &                                                                              &                        &                                                                              &                       & \textbf{5}            & \multicolumn{1}{c|}{17.90}       & \multicolumn{1}{c|}{-}             & \multicolumn{1}{c|}{19.50}       & \multicolumn{1}{c|}{1.00}        & \multicolumn{1}{c|}{-}             & \multicolumn{1}{c|}{7.40}        & \multicolumn{1}{c|}{13.70}       & \multicolumn{1}{c|}{-}             & 16.40       & \multicolumn{1}{c|}{2.40}        & \multicolumn{1}{c|}{-}             & 12.20       &                                                                                                                                                                                                    \\ \cline{7-19}
                               &                                                                              &                                                                              &                        &                                                                              &                       & 10                    & \multicolumn{1}{c|}{17.90}       & \multicolumn{1}{c|}{-}             & \multicolumn{1}{c|}{19.50}       & \multicolumn{1}{c|}{1.20}        & \multicolumn{1}{c|}{-}             & \multicolumn{1}{c|}{7.60}        & \multicolumn{1}{c|}{13.70}       & \multicolumn{1}{c|}{-}             & 16.50       & \multicolumn{1}{c|}{2.60}        & \multicolumn{1}{c|}{-}             & 11.60       &                                                                                                                                                                                                    \\ \cline{3-20} 
                               &                                                                              & \multirow{3}{*}{SS~\cite{cheng2021meta}}                                                          & \multirow{6}{*}{TCSVT 21} & \multirow{6}{*}{\textbf{50}}                                                 & \multirow{6}{*}{D}    & \textbf{1}            & \multicolumn{1}{c|}{26.90}       & \multicolumn{1}{c|}{-}             & \multicolumn{1}{c|}{25.80}       & \multicolumn{1}{c|}{0.90}        & \multicolumn{1}{c|}{-}             & \multicolumn{1}{c|}{4.20}        & \multicolumn{1}{c|}{20.40}       & \multicolumn{1}{c|}{-}             & 20.40       & \multicolumn{1}{c|}{1.50}        & \multicolumn{1}{c|}{2.30}          & 6.10        & \multirow{6}{*}{\begin{tabular}[c]{@{}p{4cm}@{}}Proposed new models, redesigning the CenterNet and incorporating a novel meta-learning method, MAML, to perform FSCIOD task\end{tabular}}                                                       \\ \cline{7-19}
                               &                                                                              &                                                                              &                        &                                                                              &                       & \textbf{5}            & \multicolumn{1}{c|}{29.20}       & \multicolumn{1}{c|}{-}             & \multicolumn{1}{c|}{27.30}       & \multicolumn{1}{c|}{1.40}        & \multicolumn{1}{c|}{-}             & \multicolumn{1}{c|}{7.10}        & \multicolumn{1}{c|}{22.30}       & \multicolumn{1}{c|}{-}             & 22.20       & \multicolumn{1}{c|}{3.10}        & \multicolumn{1}{c|}{5.50}          & 12.00       &                                                                                                                                                                                                    \\ \cline{7-19}
                               &                                                                              &                                                                              &                        &                                                                              &                       & 10                    & \multicolumn{1}{c|}{27.40}       & \multicolumn{1}{c|}{-}             & \multicolumn{1}{c|}{25.90}       & \multicolumn{1}{c|}{1.50}        & \multicolumn{1}{c|}{-}             & \multicolumn{1}{c|}{7.90}        & \multicolumn{1}{c|}{20.90}       & \multicolumn{1}{c|}{-}             & 21.40       & \multicolumn{1}{c|}{3.80}        & \multicolumn{1}{c|}{6.50}          & 13.50       &                                                                                                                                                                                                    \\ \cline{3-3} \cline{7-19}
                               &                                                                              & \multirow{3}{*}{MS~\cite{cheng2021meta}}                                                          &                        &                                                                              &                       & 1                     & \multicolumn{1}{c|}{30.70}       & \multicolumn{1}{c|}{-}             & \multicolumn{1}{c|}{27.60}       & \multicolumn{1}{c|}{1.50}        & \multicolumn{1}{c|}{-}             & \multicolumn{1}{c|}{5.50}        & \multicolumn{1}{c|}{23.40}       & \multicolumn{1}{c|}{-}             & 22.00       & \multicolumn{1}{c|}{2.50}        & \multicolumn{1}{c|}{4.50}          & 8.50        &                                                                                                                                                                                                    \\ \cline{7-19}
                               &                                                                              &                                                                              &                        &                                                                              &                       & 5                     & \multicolumn{1}{c|}{33.30}       & \multicolumn{1}{c|}{-}             & \multicolumn{1}{c|}{29.10}       & \multicolumn{1}{c|}{2.50}        & \multicolumn{1}{c|}{-}             & \multicolumn{1}{c|}{9.10}        & \multicolumn{1}{c|}{25.60}       & \multicolumn{1}{c|}{-}             & 24.10       & \multicolumn{1}{c|}{5.00}        & \multicolumn{1}{c|}{9.70}          & 14.60       &                                                                                                                                                                                                    \\ \cline{7-19}
                               &                                                                              &                                                                              &                        &                                                                              &                       & 10                    & \multicolumn{1}{c|}{31.40}       & \multicolumn{1}{c|}{-}             & \multicolumn{1}{c|}{27.80}       & \multicolumn{1}{c|}{2.60}        & \multicolumn{1}{c|}{-}             & \multicolumn{1}{c|}{9.60}        & \multicolumn{1}{c|}{24.20}       & \multicolumn{1}{c|}{-}             & 23.30       & \multicolumn{1}{c|}{6.20}        & \multicolumn{1}{c|}{11.40}         & 15.80       &                                                                                                                                                                                                    \\ \cline{2-20} 
                               & \multirow{9}{*}{\rotatebox{90}{FCOS-based}}                                                  & \multirow{3}{*}{Sylph~\cite{yin2022sylph}}                                                       & \multirow{3}{*}{CVPR 22}  & \multirow{3}{*}{50}                                                          & \multirow{3}{*}{D}    & 1                     & \multicolumn{1}{c|}{37.60}       & \multicolumn{1}{c|}{-}             & \multicolumn{1}{c|}{-}           & \multicolumn{1}{c|}{1.10}        & \multicolumn{1}{c|}{-}             & \multicolumn{1}{c|}{-}           & \multicolumn{1}{c|}{28.48}       & \multicolumn{1}{c|}{-}             & -           & \multicolumn{1}{c|}{-}           & \multicolumn{1}{c|}{-}             & -           & \multirow{3}{*}{\begin{tabular}[c]{@{}p{4cm}@{}}Introduced FCOS-based Sylph, decoupling object detection into classification and localization\end{tabular}}                                                                                     \\ \cline{7-19}
                               &                                                                              &                                                                              &                        &                                                                              &                       & 5                     & \multicolumn{1}{c|}{42.40}       & \multicolumn{1}{c|}{-}             & \multicolumn{1}{c|}{-}           & \multicolumn{1}{c|}{1.50}        & \multicolumn{1}{c|}{-}             & \multicolumn{1}{c|}{-}           & \multicolumn{1}{c|}{32.18}       & \multicolumn{1}{c|}{-}             & -           & \multicolumn{1}{c|}{-}           & \multicolumn{1}{c|}{-}             & -           &                                                                                                                                                                                                    \\ \cline{7-19}
                               &                                                                              &                                                                              &                        &                                                                              &                       & 10                    & \multicolumn{1}{c|}{42.80}       & \multicolumn{1}{c|}{-}             & \multicolumn{1}{c|}{-}           & \multicolumn{1}{c|}{1.70}        & \multicolumn{1}{c|}{-}             & \multicolumn{1}{c|}{-}           & \multicolumn{1}{c|}{32.53}       & \multicolumn{1}{c|}{-}             & -           & \multicolumn{1}{c|}{-}           & \multicolumn{1}{c|}{-}             & -           &                                                                                                                                                                                                    \\ \cline{3-20} 
                               &                                                                              & \multirow{3}{*}{MCH~\cite{feng2022incremental}}                                                         & \multirow{6}{*}{PRL 22}   & \multirow{6}{*}{50}                                                          & \multirow{6}{*}{D}    & 1                     & \multicolumn{1}{c|}{36.90}       & \multicolumn{1}{c|}{-}             & \multicolumn{1}{c|}{-}           & \multicolumn{1}{c|}{0.40}        & \multicolumn{1}{c|}{-}             & \multicolumn{1}{c|}{-}           & \multicolumn{1}{c|}{27.70}       & \multicolumn{1}{c|}{-}             & -           & \multicolumn{1}{c|}{1.00}        & \multicolumn{1}{c|}{-}             & -           & \multirow{6}{*}{\begin{tabular}[c]{@{}p{4cm}@{}}Introduced MCH and BPMCH, human memory-inspired models, outperforming ONCE by effectively transferring knowledge from base to novel classes\end{tabular}}                                       \\ \cline{7-19}
                               &                                                                              &                                                                              &                        &                                                                              &                       & 5                     & \multicolumn{1}{c|}{36.00}       & \multicolumn{1}{c|}{-}             & \multicolumn{1}{c|}{-}           & \multicolumn{1}{c|}{5.50}        & \multicolumn{1}{c|}{-}             & \multicolumn{1}{c|}{-}           & \multicolumn{1}{c|}{28.30}       & \multicolumn{1}{c|}{-}             & -           & \multicolumn{1}{c|}{14.30}       & \multicolumn{1}{c|}{-}             & -           &                                                                                                                                                                                                    \\ \cline{7-19}
                               &                                                                              &                                                                              &                        &                                                                              &                       & 10                    & \multicolumn{1}{c|}{35.50}       & \multicolumn{1}{c|}{-}             & \multicolumn{1}{c|}{-}           & \multicolumn{1}{c|}{7.80}        & \multicolumn{1}{c|}{-}             & \multicolumn{1}{c|}{-}           & \multicolumn{1}{c|}{28.60}       & \multicolumn{1}{c|}{-}             & -           & \multicolumn{1}{c|}{18.30}       & \multicolumn{1}{c|}{-}             & -           &                                                                                                                                                                                                    \\ \cline{3-3} \cline{7-19}
                               &                                                                              & \multirow{3}{*}{BPMCH~\cite{feng2022incremental}}                                                       &                        &                                                                              &                       & 1                     & \multicolumn{1}{c|}{29.40}       & \multicolumn{1}{c|}{-}             & \multicolumn{1}{c|}{-}           & \multicolumn{1}{c|}{2.40}        & \multicolumn{1}{c|}{-}             & \multicolumn{1}{c|}{-}           & \multicolumn{1}{c|}{22.60}       & \multicolumn{1}{c|}{-}             & -           & \multicolumn{1}{c|}{6.10}        & \multicolumn{1}{c|}{-}             & -           &                                                                                                                                                                                                    \\ \cline{7-19}
                               &                                                                              &                                                                              &                        &                                                                              &                       & 5                     & \multicolumn{1}{c|}{36.00}       & \multicolumn{1}{c|}{-}             & \multicolumn{1}{c|}{-}           & \multicolumn{1}{c|}{6.40}        & \multicolumn{1}{c|}{-}             & \multicolumn{1}{c|}{-}           & \multicolumn{1}{c|}{28.60}       & \multicolumn{1}{c|}{-}             & -           & \multicolumn{1}{c|}{16.40}       & \multicolumn{1}{c|}{-}             & -           &                                                                                                                                                                                                    \\ \cline{7-19}
                               &                                                                              &                                                                              &                        &                                                                              &                       & 10                    & \multicolumn{1}{c|}{35.60}       & \multicolumn{1}{c|}{-}             & \multicolumn{1}{c|}{-}           & \multicolumn{1}{c|}{7.00}        & \multicolumn{1}{c|}{-}             & \multicolumn{1}{c|}{-}           & \multicolumn{1}{c|}{28.50}       & \multicolumn{1}{c|}{-}             & -           & \multicolumn{1}{c|}{17.60}       & \multicolumn{1}{c|}{-}             & -           &                                                                                                                                                                                                    \\ \cline{2-20} 
                               & \multirow{3}{*}{\rotatebox{90}{\begin{tabular}[c]{@{}c@{}}DETR\\-based\end{tabular}}}                                                  & \multirow{3}{*}{\begin{tabular}[c]{@{}c@{}}Incremental\\-DETR~\cite{dong2023incremental}\end{tabular}} & \multirow{3}{*}{AAAI 23}  & \multirow{3}{*}{50}                                                          & \multirow{3}{*}{D}    & 1                     & \multicolumn{1}{c|}{29.40}       & \multicolumn{1}{c|}{47.10}         & \multicolumn{1}{c|}{-}           & \multicolumn{1}{c|}{1.90}        & \multicolumn{1}{c|}{2.70}          & \multicolumn{1}{c|}{-}           & \multicolumn{1}{c|}{22.50}       & \multicolumn{1}{c|}{36.00}         & -           & \multicolumn{1}{c|}{4.10}        & \multicolumn{1}{c|}{6.60}          & -           & \multirow{3}{*}{\begin{tabular}[c]{@{}p{4cm}@{}}Developed Incremental-DETR  for FSCID, which uses self-supervised learning and a fine-tuning strategy\end{tabular}}                                                                             \\ \cline{7-19}
                               &                                                                              &                                                                              &                        &                                                                              &                       & 5                     & \multicolumn{1}{c|}{30.50}       & \multicolumn{1}{c|}{48.40}         & \multicolumn{1}{c|}{-}           & \multicolumn{1}{c|}{8.30}        & \multicolumn{1}{c|}{13.30}         & \multicolumn{1}{c|}{-}           & \multicolumn{1}{c|}{24.90}       & \multicolumn{1}{c|}{39.60}         & -           & \multicolumn{1}{c|}{16.60}       & \multicolumn{1}{c|}{26.30}         & -           &                                                                                                                                                                                                    \\ \cline{7-19}
                               &                                                                              &                                                                              &                        &                                                                              &                       & 10                    & \multicolumn{1}{c|}{27.30}       & \multicolumn{1}{c|}{44.00}         & \multicolumn{1}{c|}{-}           & \multicolumn{1}{c|}{14.40}        & \multicolumn{1}{c|}{22.40}         & \multicolumn{1}{c|}{-}           & \multicolumn{1}{c|}{24.10}       & \multicolumn{1}{c|}{38.60}         & -           & \multicolumn{1}{c|}{24.60}       & \multicolumn{1}{c|}{38.40}         & -           &                                                                                                                                                                                                    \\ \hline
\multirow{12}{*}{\rotatebox{90}{Anchor-based}} & \multirow{12}{*}{\rotatebox{90}{Mask RCNN-based}} & \multirow{6}{*}{iMTFA~\cite{ganea2021incremental}}                                                       & \multirow{6}{*}{CVPR 21}  & \multirow{6}{*}{50}                                                          & \multirow{3}{*}{D}    & 1                     & \multicolumn{1}{c|}{27.81}       & \multicolumn{1}{c|}{40.11}         & \multicolumn{1}{c|}{-}           & \multicolumn{1}{c|}{3.23}        & \multicolumn{1}{c|}{5.89}          & \multicolumn{1}{c|}{-}           & \multicolumn{1}{c|}{21.67}       & \multicolumn{1}{c|}{31.55}         & -           & \multicolumn{1}{c|}{-}           & \multicolumn{1}{c|}{-}             & -           & \multirow{6}{*}{\begin{tabular}[c]{@{}p{4cm}@{}}Proposed instance segmentation setting in FSCIL and introduced the first work, iMTFA, which can perform both instance segmentation and object detection\end{tabular}}                           \\ \cline{7-19}
                               &                                                                              &                                                                              &                        &                                                                              &                       & 5                     & \multicolumn{1}{c|}{24.13}       & \multicolumn{1}{c|}{33.69}         & \multicolumn{1}{c|}{-}           & \multicolumn{1}{c|}{6.07}        & \multicolumn{1}{c|}{11.15}         & \multicolumn{1}{c|}{-}           & \multicolumn{1}{c|}{19.62}       & \multicolumn{1}{c|}{28.06}         & -           & \multicolumn{1}{c|}{-}           & \multicolumn{1}{c|}{-}             & -           &                                                                                                                                                                                                    \\ \cline{7-19}
                               &                                                                              &                                                                              &                        &                                                                              &                       & 10                    & \multicolumn{1}{c|}{23.36}       & \multicolumn{1}{c|}{32.41}         & \multicolumn{1}{c|}{-}           & \multicolumn{1}{c|}{6.97}        & \multicolumn{1}{c|}{12.72}         & \multicolumn{1}{c|}{-}           & \multicolumn{1}{c|}{19.26}       & \multicolumn{1}{c|}{27.49}         & -           & \multicolumn{1}{c|}{-}           & \multicolumn{1}{c|}{-}             & -           &                                                                                                                                                                                                    \\ \cline{6-19}
                               &                                                                              &                                                                              &                        &                                                                              & \multirow{3}{*}{S}    & 1                     & \multicolumn{1}{c|}{25.90}       & \multicolumn{1}{c|}{39.28}         & \multicolumn{1}{c|}{-}           & \multicolumn{1}{c|}{2.81}        & \multicolumn{1}{c|}{4.72}          & \multicolumn{1}{c|}{-}           & \multicolumn{1}{c|}{20.13}       & \multicolumn{1}{c|}{30.64}         & -           & \multicolumn{1}{c|}{-}           & \multicolumn{1}{c|}{-}             & -           &                                                                                                                                                                                                    \\ \cline{7-19}
                               &                                                                              &                                                                              &                        &                                                                              &                       & 5                     & \multicolumn{1}{c|}{22.56}       & \multicolumn{1}{c|}{33.25}         & \multicolumn{1}{c|}{-}           & \multicolumn{1}{c|}{5.19}        & \multicolumn{1}{c|}{8.65}          & \multicolumn{1}{c|}{-}           & \multicolumn{1}{c|}{18.22}       & \multicolumn{1}{c|}{27.10}         & -           & \multicolumn{1}{c|}{-}           & \multicolumn{1}{c|}{-}             & -           &                                                                                                                                                                                                    \\ \cline{7-19}
                               &                                                                              &                                                                              &                        &                                                                              &                       & 10                    & \multicolumn{1}{c|}{21.87}       & \multicolumn{1}{c|}{32.01}         & \multicolumn{1}{c|}{-}           & \multicolumn{1}{c|}{5.88}        & \multicolumn{1}{c|}{9.81}          & \multicolumn{1}{c|}{-}           & \multicolumn{1}{c|}{17.87}       & \multicolumn{1}{c|}{26.46}         & -           & \multicolumn{1}{c|}{-}           & \multicolumn{1}{c|}{-}             & -           &                                                                                                                                                                                                    \\ \cline{3-20} 
                               &                                                                              & \multirow{6}{*}{iFS-RCNN~\cite{nguyen2022ifs}}                                                    & \multirow{6}{*}{CVPR 22}  & \multirow{6}{*}{50}                                                          & \multirow{3}{*}{D}    & 1                     & \multicolumn{1}{c|}{40.08}       & \multicolumn{1}{c|}{-}             & \multicolumn{1}{c|}{-}           & \multicolumn{1}{c|}{4.54}        & \multicolumn{1}{c|}{-}             & \multicolumn{1}{c|}{-}           & \multicolumn{1}{c|}{31.19}       & \multicolumn{1}{c|}{-}             & -           & \multicolumn{1}{c|}{-}           & \multicolumn{1}{c|}{-}             & -           & \multirow{6}{*}{\begin{tabular}[c]{@{}p{4cm}@{}}Introduced iFS-RCNN, an extension of Mask-RCNN, leveraging probit function and uncertainty-guided bounding box prediction for instance segmentation and object detection in FSCIL\end{tabular}} \\ \cline{7-19}
                               &                                                                              &                                                                              &                        &                                                                              &                       & 5                     & \multicolumn{1}{c|}{40.06}       & \multicolumn{1}{c|}{-}             & \multicolumn{1}{c|}{-}           & \multicolumn{1}{c|}{9.91}        & \multicolumn{1}{c|}{-}             & \multicolumn{1}{c|}{-}           & \multicolumn{1}{c|}{32.52}       & \multicolumn{1}{c|}{-}             & -           & \multicolumn{1}{c|}{-}           & \multicolumn{1}{c|}{-}             & -           &                                                                                                                                                                                                    \\ \cline{7-19}
                               &                                                                              &                                                                              &                        &                                                                              &                       & 10                    & \multicolumn{1}{c|}{40.05}       & \multicolumn{1}{c|}{-}             & \multicolumn{1}{c|}{-}           & \multicolumn{1}{c|}{12.55}       & \multicolumn{1}{c|}{-}             & \multicolumn{1}{c|}{-}           & \multicolumn{1}{c|}{33.02}       & \multicolumn{1}{c|}{-}             & -           & \multicolumn{1}{c|}{-}           & \multicolumn{1}{c|}{-}             & -           &                                                                                                                                                                                                    \\ \cline{6-19}
                               &                                                                              &                                                                              &                        &                                                                              & \multirow{3}{*}{S}    & 1                     & \multicolumn{1}{c|}{36.35}       & \multicolumn{1}{c|}{-}             & \multicolumn{1}{c|}{-}           & \multicolumn{1}{c|}{3.95}        & \multicolumn{1}{c|}{-}             & \multicolumn{1}{c|}{-}           & \multicolumn{1}{c|}{28.45}       & \multicolumn{1}{c|}{-}             & -           & \multicolumn{1}{c|}{-}           & \multicolumn{1}{c|}{-}             & -           &                                                                                                                                                                                                    \\ \cline{7-19}
                               &                                                                              &                                                                              &                        &                                                                              &                       & 5                     & \multicolumn{1}{c|}{36.33}       & \multicolumn{1}{c|}{-}             & \multicolumn{1}{c|}{-}           & \multicolumn{1}{c|}{8.80}        & \multicolumn{1}{c|}{-}             & \multicolumn{1}{c|}{-}           & \multicolumn{1}{c|}{28.89}       & \multicolumn{1}{c|}{-}             & -           & \multicolumn{1}{c|}{-}           & \multicolumn{1}{c|}{-}             & -           &                                                                                                                                                                                                    \\ \cline{7-19}
                               &                                                                              &                                                                              &                        &                                                                              &                       & 10                    & \multicolumn{1}{c|}{36.32}       & \multicolumn{1}{c|}{-}             & \multicolumn{1}{c|}{-}           & \multicolumn{1}{c|}{1.06}        & \multicolumn{1}{c|}{-}             & \multicolumn{1}{c|}{-}           & \multicolumn{1}{c|}{30.41}       & \multicolumn{1}{c|}{-}             & -           & \multicolumn{1}{c|}{-}           & \multicolumn{1}{c|}{-}             & -           &                                                                                                                                                                                                    \\ \hline
\end{tabular}
\label{tab4}
\end{table*}

Given that the FSCIOD evaluation is usually conducted under different sample shots, we analyze and summarize based on the overall performance of relevant methods. It can be found from Tab.~\ref{tab4}, the top three performance methods for object detection achieved on base classes are Sylph~\cite{yin2022sylph}, iFS-RCNN~\cite{nguyen2022ifs}, and MCH~\cite{feng2022incremental}. The top three performance methods for novel COCO classes are iFS-RCNN, Incremental-DETR~\cite{dong2023incremental}, and iMTFA~\cite{ganea2021incremental}. The top three methods for overall performance on COCO are iFS-RCNN, Sylph, and MCH. Among all methods for cross-dataset evaluation on VOC, the top three performers are: Incremental-DETR, BPMCH~\cite{feng2022incremental}, and MCH. Therefore, it can be seen that iFS-RCNN based on complex Mask RCNN yields the best results, and Sylph and MCH, which are based on simple FCOS, also show good performance. In instance segmentation, only anchor-based methods have conducted the relevant evaluation, among which iMTFA has the overall best incremental segmentation ability on novel classes, but iFS-RCNN performs best on base classes. In summary, anchor-based methods are suitable for object detection and instance segmentation scenarios, with excellent performance but more complex structures; anchor-free methods are suitable for application scenarios requiring lower framework complexity and can achieve performance slightly inferior to anchor-based methods.

\subsubsection{Main Issues and Facts}
The current FSCIOD mainly faces the issue of insufficient research. In addition, the performance of current research is relatively poor compared to supervised learning methods, especially in detecting novel classes, which is far from the level of practical application. Furthermore, similar to FSCIC, FSCIOD also faces the problem of a need for more suitable evaluation metrics. The evaluation metrics used by different works vary slightly and are not yet unified.

\section{Conclusion and Outlooks}
\label{sec:CO}

In this paper, we present a comprehensive and systemic survey of FSCIL, covering its background and significance, problem definition, core challenges, general schemes, relations with related problems, datasets, evaluation protocols, and metrics. We focused on the classification and object detection tasks in FSCIL, summarized the relevant works, analyzed their performance, and summarized the main issues and facts faced by FSCIL. Considering that FSCIL is still in its infancy, we attempt to offer valuable insights and discuss potential directions.

\subsection{Human-machine Gap in FSCIL}

The memory learning in the human brain can be categorized into three main processes: encoding, storage, and retrieval~\cite{klein2015memory}. In the encoding phase, the brain efficiently processes information through associative learning and abstract thinking, effectively encoding features of new categories even with limited samples. During the storage phase, the hippocampus converts short-term memories into long-term memories, forming stable neural networks across different regions of the cerebral cortex. In the retrieval phase, existing memories may be consolidated, updated, or actively forgotten in conjunction with new information, leading to the formation of memories adapted to the current environment. This sequence of processes highlights the brain's efficient knowledge handling capabilities.

Currently, some IL research, such as the method proposed by \emph{ZKudithipudi et al.}~\cite{kudithipudi2022biological} that emulates the Drosophila's mushroom body's mechanisms, attempts to enhance model memory capabilities by bio-inspired intelligence. However, a systematic bio-inspired approach in FSCIL is still lacking. Current FSCIL models lack associative learning and abstract thinking in limited sample learning, and there is room for improvement in prior knowledge acquisition. These models typically use one model for storing all knowledge from continual learning, suggesting the need for exploring multi-modular knowledge storage and long-short term memory mechanisms. Additionally, FSCIL requires proactive strategies for knowledge consolidation, updating, and personalized management, such as actively forgetting infrequent knowledge, reinforcing challenging knowledge, and integrating consistent knowledge.

\subsection{Practical Settings in FSCIL}
The current FSCIL setting, based on \emph{Tao et al.}~\cite{tao2020few}, is idealistic. The real world requires practical settings. Some research has improved the FSCIL setting to better adapt to the real-world, for example: (a) \emph{FSCIL with limited base samples}: Ensuring that the base session has abundant samples is challenging in some situations. Thus, \emph{Kalla and Biswas}~\cite{kalla2022s3c} suggested the FSCIL-lb setting with fewer required base training samples; (b) \emph{FSCIL with imbalanced sessions}: Considering the practical difficulty in ensuring the $N-$way $K-$shot format, \emph{Kalla and Biswas}~\cite{kalla2022s3c} proposed the FSCIL-im setting, where the incremental sessions appear with an imbalanced data distribution; (c) \emph{Semi-supervised FSCIL}: Some scenarios have some available unlabeled data. \emph{Cui et al.}~\cite{cui2021semi} leveraged them to propose semi-supervised FSCIL. 

Despite some efforts to propose settings that better match real-world situations, some directions are still worth exploring: (a) \emph{Cross-domain FSCIL}: Considering the domain changes in the real world (e.g., changes in imaging condition and environment), FSCIL should be robust under cross-domain conditions; (b) \emph{FSCIL with repetition}: The no repetition constraint of current FSCIL doesn't reflect practical scenarios where class recurrence is common. Researching how to utilize these repetitions (considering that current samples may be scarce, but could increase in the future) can improve the practicality; (c) \emph{Incomplete FSCIL}: In real-world scenarios, where most classes have ample training samples but some are scarce, the assumption of uniformly few-shot data is unrealistic. Hence, investigating incomplete FSCIL, encompassing incremental sessions with classes of varying sample availability, is also meaningful; (d) \emph{Federated FSCIL}: Combining the privacy and distributed features of federated learning with FSCIL's ability to learn from limited data, this method aims to create models that are privacy-aware and adaptable to multiple clients with limited and dynamic data.

\subsection{Knowledge Acquisition and Update in FSCIL}
FSCIL involves a continuous learning process with base and incremental sessions, so knowledge acquisition and update are analyzed in two parts:

\textbf{Base Stage:} Effective initialization of the backbone is crucial for ensuring base class performance and generalization for future incremental classes. Current methods often rely on a large number of base class samples for backbone network initialization, which may not align with reality and whose generalization capabilities are difficult to accurately assess. To enhance generalization, researchers have tried introducing strategies like self-supervised learning and forward compatibility, but these usually depend on sufficient base class data. There is a lack of research on initial knowledge acquisition without specific requirements for the base data. Therefore, exploring methods to enrich initial stage knowledge acquisition is important. From the data perspective, increasing data diversity and improving knowledge learning strategies, such as exploring data augmentation, data generation, introducing unsupervised data, and optimizing backbone learning methods, are essential. Additionally, introducing pre-trained models and other prior knowledge can be considered. For instance, foundation models like CLIP, SAM, and GPT, which combine self-supervised or semi-supervised pre-training with prompt engineering, have shown excellent generalization and transfer capabilities, offering new possibilities for enhancing FSCIL model performance. Some recent works have attempted to incorporate foundation models to address FSCIL challenges. For example, \emph{D'Alessandro et al.}~\cite{d2023multimodal} designed a prompt learning strategy for CLIP in FSCIL, and \emph{Zhang et al.}~\cite{zhang2024few} leveraged the RETFound foundation model to enhance feature learning in few-shot class-incremental retinal disease recognition. However, these approaches have not yet become mainstream, and attention to fairness in experimental comparisons is still necessary.

\textbf{Incremental Stage:} 
In incremental sessions, models typically initialize with weights from prior phases, focusing on learning new classes and preserving existing knowledge. Challenges arise from limited new samples and restricted access to complete old data, making effective learning of new categories and old knowledge retention pivotal. Current solutions include freezing the backbone network and using class prototype averaging, demanding robust generalization and discrimination from the network, yet possibly leading to reduced performance as new classes increase. An alternative is maintaining key parameters for new class learning, though this risks diminishing old class performance and complicates parameter evaluation due to deep learning models' opaque nature. KD is also commonly used, but how to effectively learn new categories and select efficient old samples for distillation is still a direction to be further explored.

\subsection{Applications and Safety in FSCIL}

\textbf{Application Scenarios:}
Current FSCIL research mainly targets image classification, with emerging yet non-systematic studies in visual object detection, natural language processing, lip reading, remote sensing, and robotics. Most work evaluates performance on benchmark datasets, with real-world applications still evolving. In many application scenarios, the demand for the few-shot continuous learning capability is significant. For instance, in applications like video analysis, service robotics in hotels, and autonomous driving, the need for FSCIL technology is evident. These fields often require learning new classes from limited data, maintaining high accuracy with scarce samples, and adapting to new categories in dynamic environments, underscoring FSCIL's importance and potential.

\textbf{Privacy and Safety:}
Privacy and Security: Privacy protection is a key issue in the application of FSCIL. To address catastrophic forgetting, some FSCIL studies store old category samples for replay, which could lead to privacy breaches when dealing with tasks involving private data. Currently, research on privacy protection in FSCIL is relatively limited, especially in the context of the increasing prevalence of deep learning technologies. Despite improvements in FSCIL's accuracy, AI systems based on deep learning are susceptible to security threats like adversarial and data poisoning attacks. Therefore, in-depth research into the security and privacy protection aspects of FSCIL is essential for its widespread application across various scenarios.

\footnotesize
\bibliographystyle{IEEEtran}
\bibliography{IEEEabrv,myreference}

\vspace{-35pt}
\begin{IEEEbiography}[{\includegraphics[width=1in,height=1.25in,clip,keepaspectratio]{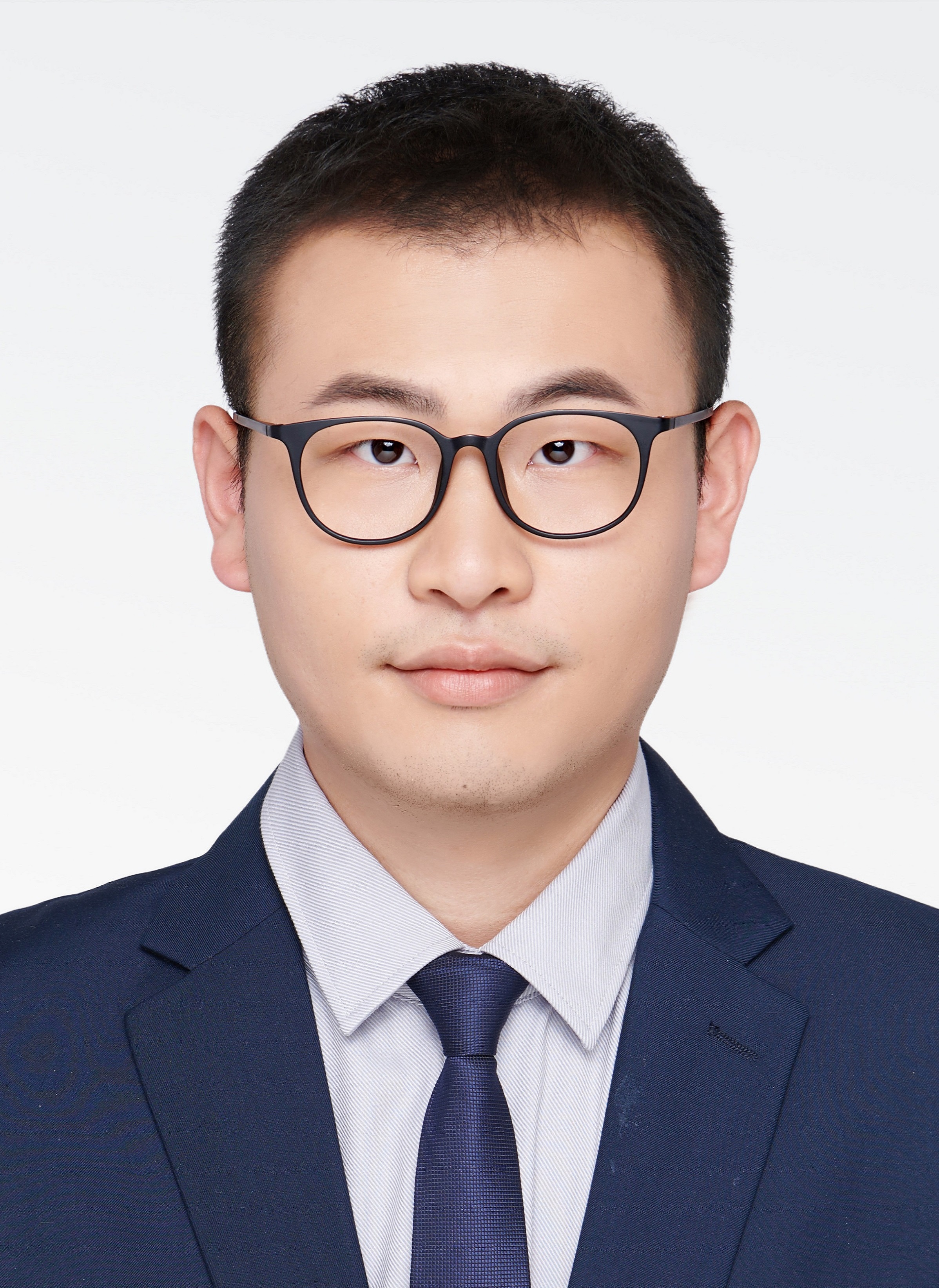}}]{Jinghua Zhang} received the B.E. degree from the Hefei University, China, in 2018, and the M.E. degree from the Northeastern University, China, in 2021. He is currently pursuing the Ph.D. degree in control science and engineering with the National University of Defense Technology, and is also with the Center for Machine Vision and Signal Analysis, University of Oulu. His research interests include computer vision and deep learning.
\end{IEEEbiography}
\vspace{-40pt}
\begin{IEEEbiography}[{\includegraphics[width=1in,height=1.25in,clip,keepaspectratio]{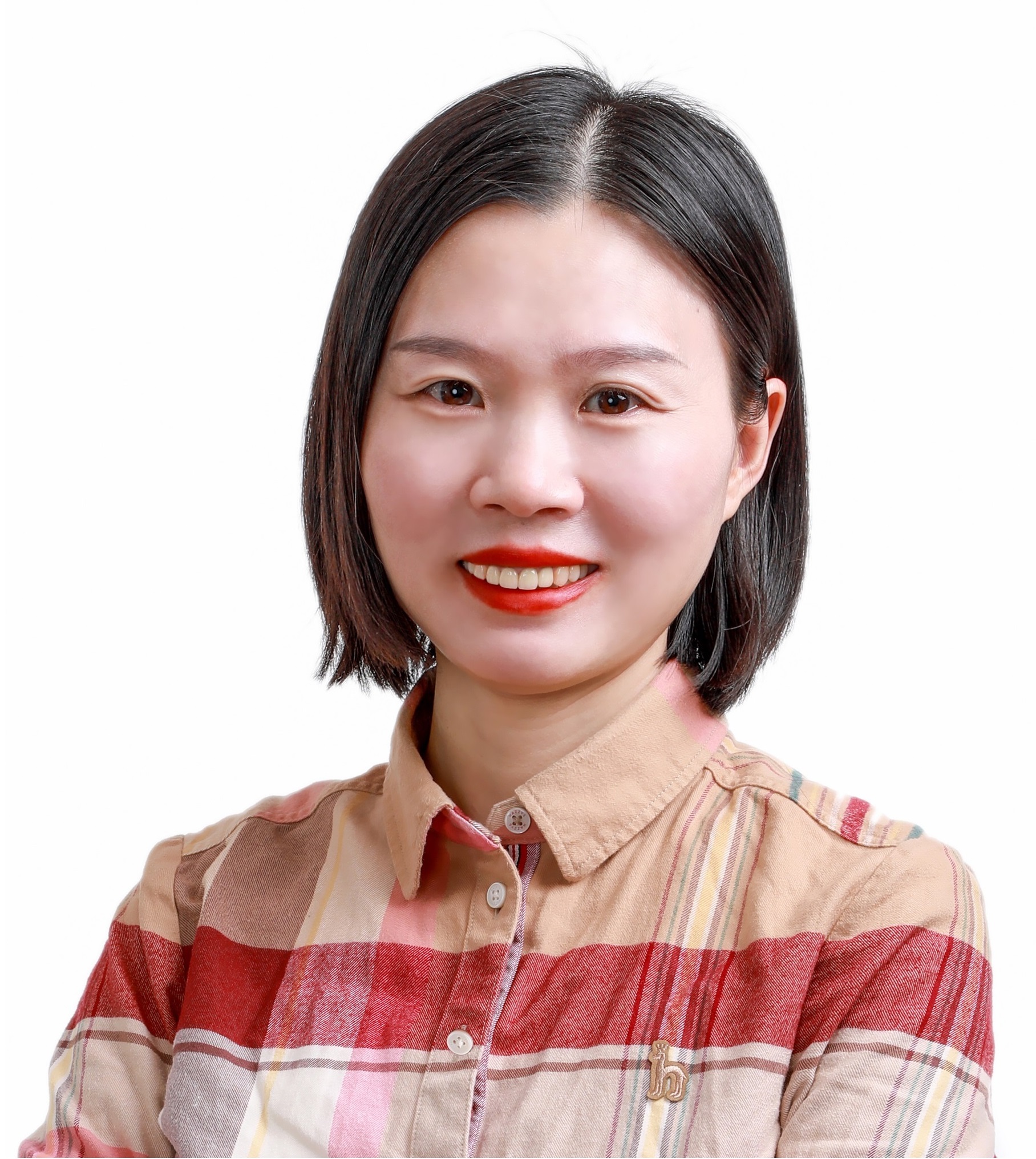}}]{Li Liu} received her Ph.D. from National University of Defense Technology, China, in 2012 and is now a Full Professor there. She has visited the University of Waterloo, Chinese University of Hong Kong, and University of Oulu. She has co-chaired workshops for CVPR and ICCV, served as lead guest editor for IEEE TPAMI and IJCV, and is an Associate Editor for IEEE TCSVT and Pattern Recognition. Her research in computer vision, pattern recognition, and machine learning has garnered over 16,000 citations.
\end{IEEEbiography}
\vspace{-40pt}
\begin{IEEEbiography}[{\includegraphics[width=1in,height=1.25in,clip,keepaspectratio]{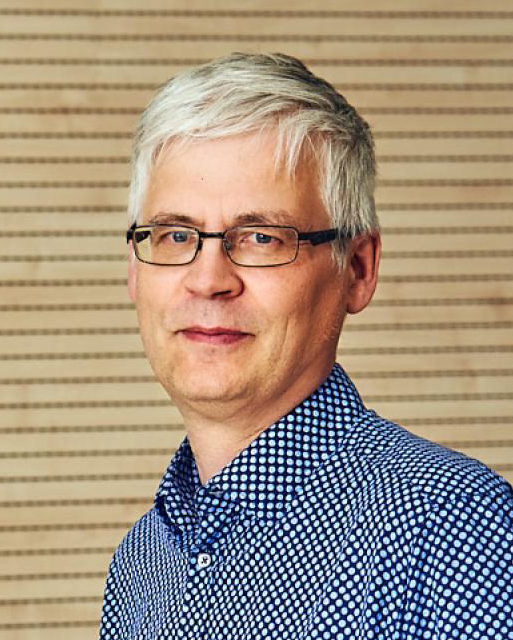}}]{Olli Silvén} received the M.Sc. and Ph.D. degrees in electrical and computer engineering from the University of Oulu, Finland, in 1982 and 1988, respectively. Since 1996, he has been a professor of signal processing engineering with the University of Oulu. He has contributed to the development of numerous solutions from real-time 3-D imaging in reverse vending machines to IP blocks for mobile video coding. His research focuses on ultra-energy-efficient-embedded signal processing and machine vision system design.
\end{IEEEbiography}
\vspace{-40pt}
\begin{IEEEbiography}[{\includegraphics[width=1in,height=1.25in,clip,keepaspectratio]{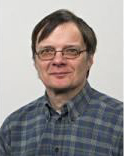}}]{Matti Pietikäinen}, who earned his Ph.D. degree from the University of Oulu and serves as an emeritus professor at its Center for Machine Vision and Signal Analysis, is notable for his contributions to Local Binary Pattern (LBP). His work has attracted about 91,600 citations on Google Scholar. He has been honored with the Koenderink Prize in 2014 and the IAPR King-Sun Fu Prize in 2018 for his machine vision contributions. He is also an IEEE fellow, recognized for his work in machine vision.
\end{IEEEbiography}
\vspace{-40pt}
\begin{IEEEbiography}[{\includegraphics[width=1in,height=1.25in,clip,keepaspectratio]{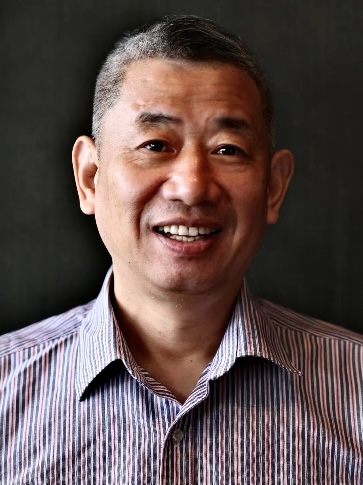}}]{Dewen Hu} received his B.S. and M.S. degrees from Xi’an Jiaotong University, China, in 1983 and 1986, and his Ph.D. from the National University of Defense Technology, China, in 1999. He is currently a Professor at the same university. He has visited the University of Sheffield, U.K., in 1995-1996. He has over 400 publications in journals and conferences like PNAS, IEEE TPAMI, and IJCV, focusing on pattern recognition and cognitive science, and serves as an associate editor for IEEE TSMCS.
\end{IEEEbiography}

 \newpage
\end{document}